\documentclass{article}

\PassOptionsToPackage{numbers, compress}{natbib}
\usepackage[final]{neurips_data_2022}

\usepackage[utf8]{inputenc} %
\usepackage[T1]{fontenc}    %
\usepackage{hyperref}       %
\usepackage{url}            %
\usepackage{booktabs}       %
\usepackage{amsfonts}       %
\usepackage{amsmath}
\usepackage{nicefrac}       %
\usepackage{microtype}      %
\usepackage[usenames,dvipsnames]{xcolor}
\usepackage{xspace}
\usepackage{graphicx}
\usepackage{caption}
\usepackage{subcaption}
\usepackage{enumitem}
\usepackage{multirow, makecell}
\usepackage{soul}

\usepackage{multicol}
\usepackage{array}
\usepackage{amssymb,bm,amsthm}
\usepackage[graphicx]{realboxes}
\usepackage[ruled,linesnumbered]{algorithm2e}
\usepackage{bbm}
\usepackage{listings}

\usepackage{tabularx}
\newcolumntype{L}[1]{>{\raggedright\let\newline\\\arraybackslash\hspace{0pt}}m{#1}}
\newcolumntype{C}[1]{>{\centering\let\newline\\\arraybackslash\hspace{0pt}}m{#1}}
\newcolumntype{R}[1]{>{\raggedleft\let\newline\\\arraybackslash\hspace{0pt}}m{#1}}
\newcolumntype{Y}{>{\centering\arraybackslash}X}
\newcommand{\addRoboFig}[1]{\includegraphics[width=\linewidth]{tables/imgs/#1.png}}

\usepackage{soul}

\DeclareRobustCommand{\Rebuttal}[1]{#1}

\definecolor{snsgray}{RGB}{179, 179, 179}
\definecolor{snsorange}{RGB}{252, 141, 98}
\definecolor{snsblue}{RGB}{141, 160, 203}

\definecolor{coolgrey}{RGB}{157,157,157}
\definecolor{lightgrey}{RGB}{235,238,238}
\definecolor{lightteal}{RGB}{198,211,222}
\definecolor{cyan}{RGB}{136, 204, 238}
\definecolor{teal}{RGB}{68, 170, 153}
\definecolor{sand}{RGB}{221, 204, 119}
\definecolor{rose}{RGB}{204, 102, 119}
\definecolor{red}{RGB}{250, 94, 91}
\definecolor{orange}{RGB}{255, 200, 63}
\definecolor{yellow}{RGB}{254, 239, 109}

\definecolor{darkgreen}{rgb}{0.09, 0.45, 0.27}

\newcommand{\minedojo}[0]{\mbox{\textsc{MineDojo}}\xspace}
\newcommand{\mineclip}[0]{\textsc{MineCLIP}\xspace}
\newcommand{\mineclipavg}[0]{\textsc{MineCLIP}{\small [avg]}\xspace}
\newcommand{\mineclipattn}[0]{\textsc{MineCLIP}{\small [attn]}\xspace}

\newcommand{\openaiclip}[0]{CLIP\textsubscript{OpenAI}\xspace}

\newcommand{\bestscore}[1]{\textcolor{darkgreen}{\mathbf{#1}}}

\newcommand{\para}[1]{\paragraph{#1}\looseness=-1}

\title{\minedojo: Building Open-Ended \\ Embodied Agents with Internet-Scale Knowledge}

\newcommand{\weburl}{\url{https://minedojo.org}}

\author{%
    Linxi Fan$^1$, Guanzhi Wang$^{2*}$, Yunfan Jiang$^{3*}$, Ajay Mandlekar$^1$, Yuncong Yang$^4$,\\
    \textbf{Haoyi Zhu$^5$, Andrew Tang$^4$, De-An Huang$^1$, Yuke Zhu$^{1\,6 \dagger}$, Anima Anandkumar$^{1\,2 \dagger}$}\\
    $^1$NVIDIA, $^2$Caltech, $^3$Stanford, $^4$Columbia, $^5$SJTU, $^6$UT Austin\\
    $^*$Equal contribution \, $^\dagger$Equal advising \\
    \weburl
}

\begin{document}

\maketitle

\begin{abstract}

Autonomous agents have made great strides in specialist domains like Atari games and Go. However, they typically learn \textit{tabula rasa} in isolated environments with limited and manually conceived objectives, thus failing to generalize across a wide spectrum of tasks and capabilities. 
Inspired by how humans continually learn and adapt in the open world, we advocate a trinity of ingredients for building generalist agents: 1) an environment that supports a multitude of tasks and goals, 2) a large-scale database of multimodal knowledge, and 3) a flexible and scalable agent architecture.  
We introduce \minedojo, a new framework built on the popular \textit{Minecraft} game that features a simulation suite with thousands of diverse open-ended tasks and an internet-scale knowledge base with Minecraft videos, tutorials, wiki pages, and forum discussions. Using \minedojo's data, we propose a novel agent learning algorithm that leverages large pre-trained video-language models as a learned reward function. Our agent is able to solve a variety of open-ended tasks specified in free-form language without any manually designed dense shaping reward. We open-source the simulation suite, knowledge bases, algorithm implementation, and pretrained models (\weburl) to promote research towards the goal of generally capable embodied agents.

\end{abstract}

\section{Introduction}
\label{sec:introduction}
\begin{figure}
    \centering
    \includegraphics[width=1.0\linewidth]{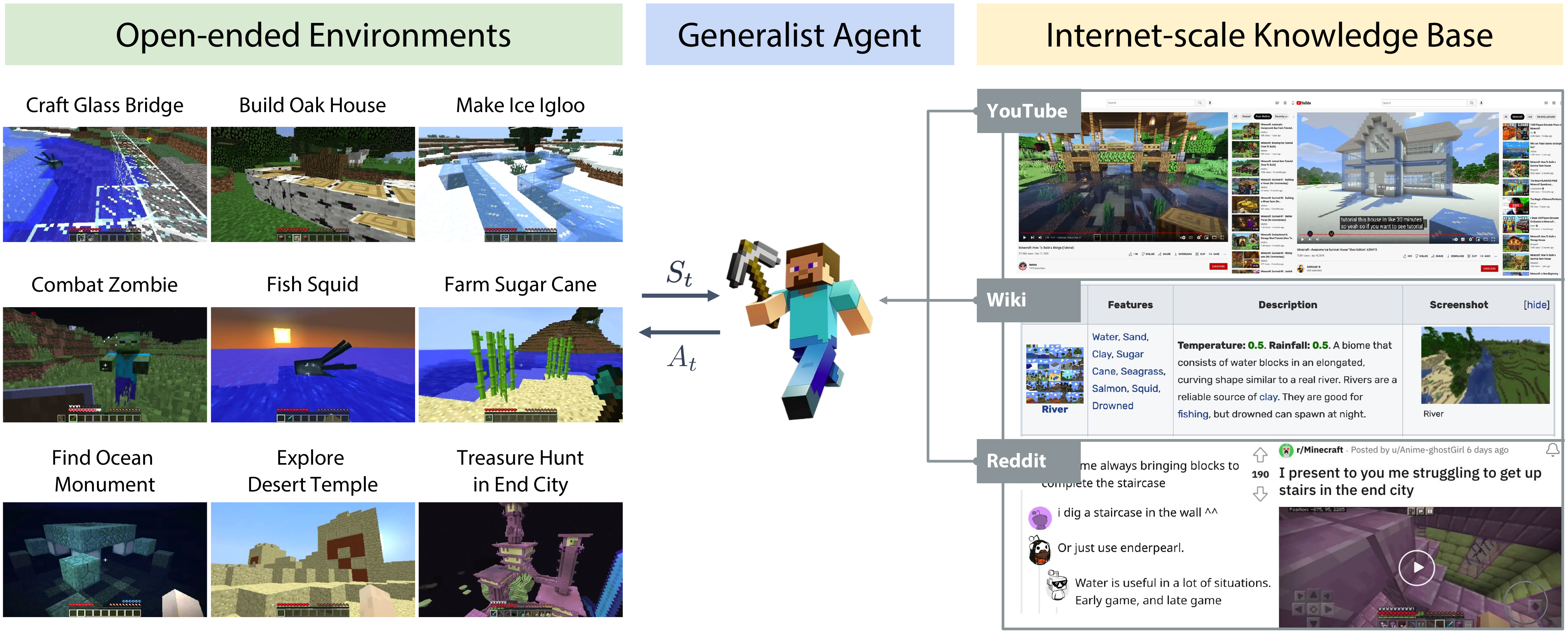}
    \caption{
    \minedojo is a novel framework for developing open-ended, generally capable agents that can learn and adapt continually to new goals. \minedojo features a benchmarking suite with \textbf{\textit{thousands} of diverse open-ended tasks} specified in natural language prompts, and also provides an \textbf{internet-scale, multimodal knowledge base} of YouTube videos, Wiki pages, and Reddit posts. The database captures the collective experience and wisdom of millions of Minecraft gamers for an AI agent to learn from. Best viewed zoomed in.}
    \label{fig:pull_figure}
\end{figure}
Developing autonomous embodied agents that can attain human-level performance across a wide spectrum of tasks has been a long-standing goal for AI research. There has been impressive progress towards this goal, most notably in games \cite{mnih2013playing, openai2019dota, vinyals2019alphastar} and robotics \cite{kolve2017ai2thor, savva2019habitat, zhu2020robosuite, xia2019igibson0.5, shen2020igibson}. These embodied agents are typically trained \textit{tabula rasa} in isolated worlds with limited complexity and diversity. Although highly performant, they are specialist models that do not generalize beyond a narrow set of tasks. 
In contrast, humans inhabit an infinitely rich reality, continuously learn from and adapt to a wide variety of open-ended tasks, and are able to leverage large amount of prior knowledge from their own experiences as well as others.  

We argue that \textbf{three main pillars} are necessary for generalist embodied agents to emerge. First, the environment in which the agent acts needs to \textbf{enable an unlimited variety of open-ended goals} \cite{standish2003open, langdon2005pfeiffer, taylor2016open, stanley2017open}. Natural evolution is able to nurture an ever-expanding tree of diverse life forms thanks to the infinitely varied ecological settings that the Earth supports \cite{stanley2017open, wang2019paired}. This process has not stagnated for billions of years. In contrast, today's agent training algorithms cease to make new progress after convergence in narrow environments \cite{mnih2013playing,zhu2020robosuite}.
Second, a \textbf{large-scale database of prior knowledge} is necessary to facilitate learning in open-ended settings. Just as humans frequently learn from the internet, agents should also be able to harvest practical knowledge encoded in large amounts of video demos~\cite{goyal2017something, miech2019howto100m}, multimedia tutorials~\cite{minecraftwiki}, and forum discussions~\cite{volske-etal-2017-tl, DBLP:conf/naacl/KimKK19, henderson2019repository}. In a complex world, it would be extremely inefficient for an agent to learn everything from scratch through trial and error.  
Third, the \textbf{agent's architecture} needs to be flexible enough to pursue any task in open-ended environments, and scalable enough to convert large-scale knowledge sources into actionable insights~\cite{ranzato2021decisiontransformer,reid2022wikipedia}. This motivates the design of an agent that has a unified observation/action space, conditions on natural language task prompts, and adopts the Transformer pre-training paradigm~\cite{devlin2018bert,radford2019gpt2, brown2020gpt3} to internalize knowledge effectively. 

In light of these three pillars, we introduce \minedojo, a new framework to help the community develop open-ended, generally-capable agents. It is built on the popular Minecraft game, where a player explores a procedurally generated 3D world with diverse types of terrains to roam, materials to mine, tools to craft, structures to build, and wonders to discover. 
Unlike most other games \cite{mnih2013playing, openai2019dota, vinyals2019alphastar}, Minecraft defines no specific reward to maximize and no fixed storyline to follow, making it well suited for developing open-ended environments for embodied AI research. 
We make the following three major contributions: 

\para{1. Simulation platform with thousands of diverse open-ended tasks.}\minedojo provides convenient APIs on top of Minecraft that standardize task specification, world settings, and agent's observation/action spaces. We introduce a benchmark suite that consists of thousands of natural language-prompted tasks, making it \textit{two orders of magnitude} larger than prior Minecraft benchmarks like the MineRL Challenge \cite{guss2019minerl,kanervisto2022minerl}. 
The suite includes long-horizon, open-ended tasks that cannot be easily evaluated through automated procedures, such as ``\textit{build an epic modern house with two floors and a swimming pool}''. %
Inspired by the Inception score~\cite{DBLP:conf/nips/SalimansGZCRCC16} and FID score~\cite{DBLP:conf/nips/HeuselRUNH17} that are commonly used to assess AI-generated image quality, we introduce a novel agent evaluation protocol using a large video-language model pre-trained on Minecraft YouTube videos. 
This complements human scoring \cite{shah2021basalt} that is precise but more expensive. 
Our learned evaluation metric has good agreement with human judgment in a subset of the full task suite considered in the experiments.

\para{2. Internet-scale multimodal Minecraft knowledge base.} Minecraft has more than 100 million active players \cite{enwiki:1092238294}, who have collectively generated an enormous wealth of data. They record tutorial videos, stream live play sessions, compile recipes, and discuss tips and tricks on forums. \minedojo features a massive collection of 730K+ YouTube videos with time-aligned transcripts, 6K+ free-form Wiki pages, and 340K+ Reddit posts with multimedia contents (Fig. \ref{fig:dataset}). We hope that this enormous knowledge base can help the agent acquire diverse skills, develop complex strategies, discover interesting objectives,  and learn actionable representations automatically.

\para{3. Novel algorithm for embodied agents with large-scale pre-training.} 

\begin{figure*}[t!]
    \centering
    \includegraphics[width=\linewidth]{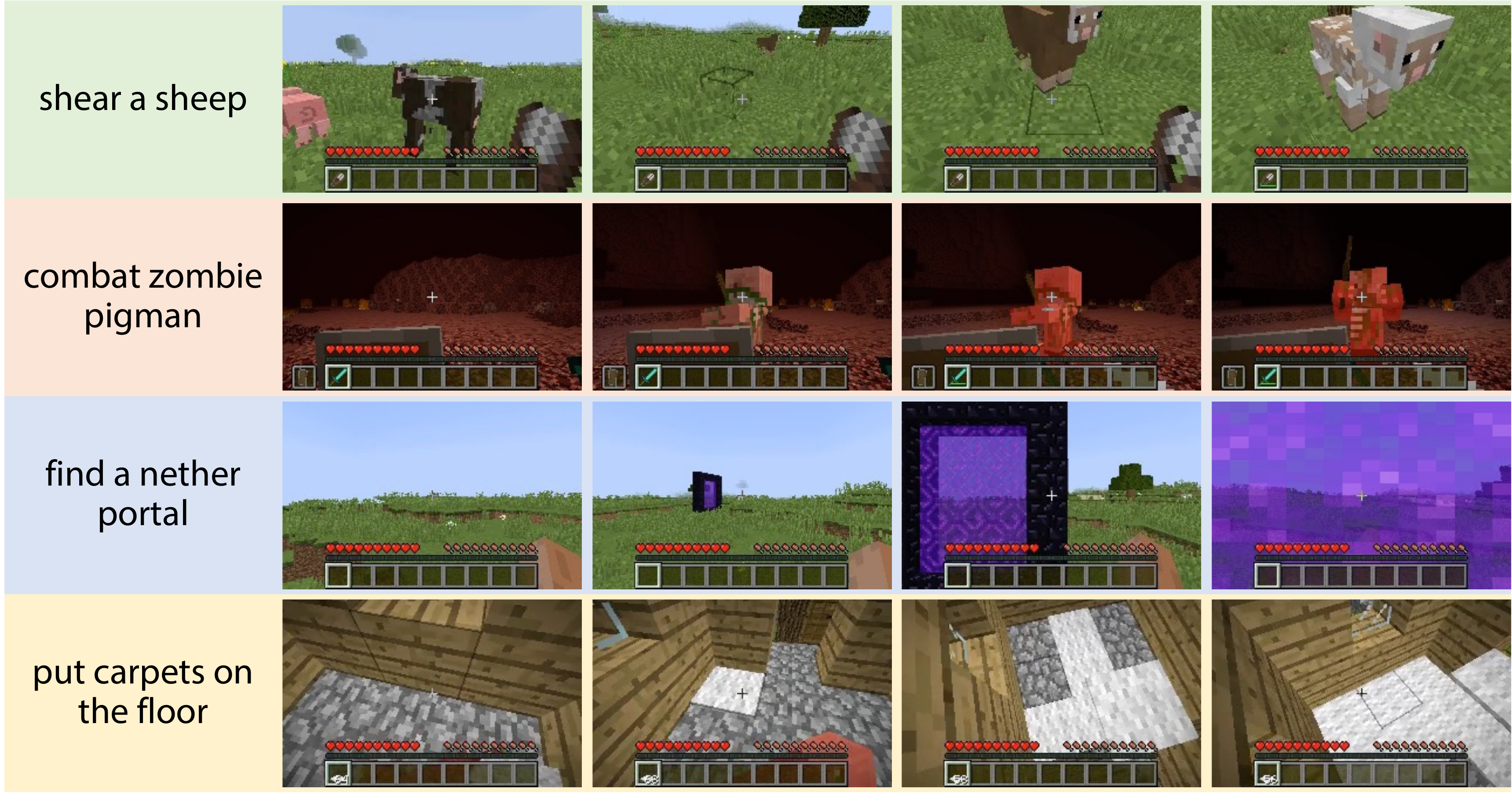}
    \caption{Visualization of our agent's learned behaviors on four selected tasks. Leftmost texts are the task prompts used in training. Best viewed on a color display.}
    \label{fig:behavior_visualization}
\end{figure*}

We develop a new learning algorithm for embodied agents that makes use of the internet-scale domain knowledge we have collected from the web. Using the massive volume of YouTube videos from \minedojo, we train a video-text contrastive model in the spirit of CLIP \cite{radford2021clip}, which associates natural language subtitles with their time-aligned video segments. 
We demonstrate that this learned correlation score can be used effectively as an \textit{open-vocabulary, massively multi-task reward function} for RL training. Our agent solves the majority of 12 tasks in our experiment using the learned reward model (Fig.~\ref{fig:behavior_visualization}). It achieves competitive performance to agents trained with meticulously engineered dense-shaping rewards, and in some cases outperforms them, with up to 73\% improvement in success rates. For open-ended tasks that do not have a simple success criterion, our agents also perform well without any special modifications.

In summary, this paper proposes an open-ended task suite, internet-scale domain knowledge, and agent learning with recent advances on large pre-trained models \cite{bommasani2021foundation}. \Rebuttal{We have open-sourced \minedojo's simulator, knowledge bases, algorithm implementations, pretrained model checkpoints, and task curation tools at \mbox{\url{https://minedojo.org/}}.} We hope that \minedojo will serve as an effective starter framework for the community to develop new algorithms and advance towards generally capable embodied agents.

\section{\minedojo Simulator \& Benchmark Suite}
\label{sec:benchmark}
\minedojo offers a set of simulator APIs help researchers develop generally capable, open-ended agents in Minecraft. 
It builds upon the open-source MineRL codebase \cite{guss2019minerl} and makes the following upgrades: 
1) We provide \textbf{unified observation and action spaces} across all tasks, facilitating the development of multi-task and continually learning agents that can constantly adapt to new scenarios and novel tasks. This deviates from the MineRL Challenge design that tailors observation and action spaces to individual tasks;
2) Our simulation unlocks all three types of worlds in Minecraft, including the \textit{Overworld}, the \textit{Nether}, and the \textit{End}, which \textbf{substantially expands the possible task space}, while MineRL only supports the Overworld natively; and 3) We provide convenient APIs to configure initial conditions and world settings to standardize our tasks. 

With this \minedojo simulator, we define thousands of benchmarking tasks, which are divided into two categories: 1) \textit{Programmatic tasks} that can be automatically assessed based on the ground-truth simulator states; 
and 2) \textit{Creative tasks} that do not have well-defined or easily-automated success criteria, which motivates our novel evaluation protocol using a learned model (Sec. \ref{sec:method}). 
To scale up the number of Creative tasks, we mine ideas from YouTube tutorials and use OpenAI's GPT-3~\cite{brown2020gpt3} service to generate substantially more task definitions. 
Compared to Creative tasks, Programmatic tasks are simpler to get started, but tend to have restricted scope, limited language variations, and less open-endedness in general.

\subsection{Task Suite I: Programmatic Tasks}
\label{sec:benchmark-programmatic}
We formalize each programmatic task as a 5-tuple: $T = (G, \mathcal{G}, \mathcal{I}, f_\mathcal{S}, f_\mathcal{R})$. 
$G$ is an English description of the task goal, such as ``\textit{find material and craft a gold pickaxe}''. 
$\mathcal{G}$ is a natural language guidance that provides helpful hints, recipes, or advice to the agent.
We leverage OpenAI's \texttt{GPT-3-davinci} API to automatically generate detailed guidance for a subset of the tasks.
For the example goal ``\textit{bring a pig into Nether}'', GPT-3 returns: \texttt{1) Find a pig in the overworld;
2) Right-click on the pig with a lead;
3) Right-click on the Nether Portal with the lead and pig selected;
4) The pig will be pulled through the portal!}
$\mathcal{I}$ is the initial conditions of the agent and the world, such as the initial inventory, spawn terrain, and weather.
$f_\mathcal{S}$: $s_t \rightarrow \{0, 1\}$ is the success criterion, a deterministic function that maps the current world state $s_t$ to a Boolean success label. 
$f_\mathcal{R}$: $s_t \rightarrow \mathbb{R}$ is an optional dense reward function. We only provide $f_\mathcal{R}$ for a small subset of the tasks in \minedojo due to the high costs of meticulously crafting dense rewards. \Rebuttal{For our current agent implementation (\mbox{Sec. \ref{sec:method-mineclip-pretraining}}), we do not use detailed guidance. Inspired by concurrent works SayCan~\mbox{\cite{ahn2022saycan}} and Socratic Models~\mbox{\cite{zeng2022socratic}}, one potential idea is to feed each step in the guidance to our learned reward model sequentially so that it becomes a stagewise reward function for a complex multi-stage task.}

\minedojo provides 4 categories of programmatic tasks with 1,581 template-generated natural language goals to evaluate the agent's different capabilities systematically and comprehensively: 

\begin{enumerate}[leftmargin=3em]
    \item \textbf{Survival}: surviving for a designated number of days.
    \item \textbf{Harvest}: finding, obtaining, cultivating, or manufacturing hundreds of materials and objects.
    \item \textbf{Tech Tree}: crafting and using a hierarchy of tools.
    \item \textbf{Combat}: fighting various monsters and creatures that require fast reflex and martial skills.
\end{enumerate}

Each task template has a number of variations based on the terrain, initial inventory, quantity, etc., which form a flexible spectrum of difficulty. In comparison, the NeurIPS MineRL Diamond challenge \cite{guss2019minerl} is a subset of our programmatic task suite, defined by the task goal ``\textit{obtain 1 diamond}" in \minedojo.  

\subsection{Task Suite II: Creative Tasks}
\label{sec:benchmark-creative}

We define each creative task as a 3-tuple, $T = (G, \mathcal{G}, \mathcal{I})$, which differs from programmatic tasks due to the lack of straightforward success criteria. 
Inspired by model-based metrics like the Inception score~\cite{DBLP:conf/nips/SalimansGZCRCC16} and FID score~\cite{DBLP:conf/nips/HeuselRUNH17} for image generation, we design a novel task evaluation metric based on a pre-trained contrastive video-language model (Sec. \ref{sec:method-mineclip-pretraining}). In the experiments, we find that the learned metric exhibits a high level of agreement with human evaluations (see Table \ref{table:eval-metric-quality}).  

We brainstorm and author 216 Creative tasks, such as ``\textit{build a haunted house with zombie inside}'' and ``\textit{race by riding a pig}''. Nonetheless, such a manual approach is not scalable. Therefore, we develop two systematic approaches to extend the total number of task definitions to 1,560. This makes our Creative tasks \textit{3 orders of magnitude} larger than Minecraft BASALT challenge \cite{shah2021basalt}, which has 4 Creative tasks.

\para{Approach 1. Task Mining from YouTube Tutorial Videos.} We identify our YouTube dataset as a rich source of tasks, as many human players demonstrate and narrate creative missions in the tutorial playlists. To collect high-quality tasks and accompanying videos, we design a 3-stage pipeline that makes it easy to find and annotate interesting tasks (see Sec.~\ref{supp:sec:creative_tasks} for details).
Through this pipeline, we extract 1,042 task ideas from the common wisdom of a huge number of veteran Minecraft gamers, such as ``\textit{make an automated mining machine}'' and ``\textit{grow cactus up to the sky}''. 

\para{Approach 2. Task Creation by GPT-3.} We leverage GPT-3's few-shot capability to generate new task ideas by seeding it with the tasks we manually author or mine from YouTube. The prompt template is: \texttt{Here are some example creative tasks in Minecraft: \{a few examples\}. Let's brainstorm more detailed while reasonable creative tasks in Minecraft}.\\
GPT-3 contributes 302 creative tasks after de-duplication, and demonstrates a surprisingly proficient understanding of Minecraft terminology.

\subsection{Collection of Starter Tasks}

\Rebuttal{We curate a set of 64 core tasks for future researchers to get started more easily. If their agent works well on these tasks, they can more confidently scale to the full benchmark. }

\begin{itemize}[leftmargin=2em,labelsep=0.7em]
    \item \textbf{32 programmatic tasks}: 16 ``standard'' and 16 ``difficult'', spanning all 4 categories (survival, harvesting, combat, and tech tree). We rely on our Minecraft knowledge to decide the difficulty level. ``Standard'' tasks require fewer steps and lower resource dependencies to complete. 
    \item \textbf{32 creative tasks}: 16 ``standard'' and 16 ``difficult''. Similarly, tasks labeled with ``standard'' are typically short-horizon tasks. 
\end{itemize}

\Rebuttal{We recommend that researchers run 100 evaluation episodes for each task and report the percentage success rate. The programmatic tasks have ground-truth success, while the creative tasks need our novel evaluation protocol (Sec.~\mbox{\ref{sec:experiments}}). }

\begin{figure}
    \centering
    \includegraphics[width=1.0\linewidth]{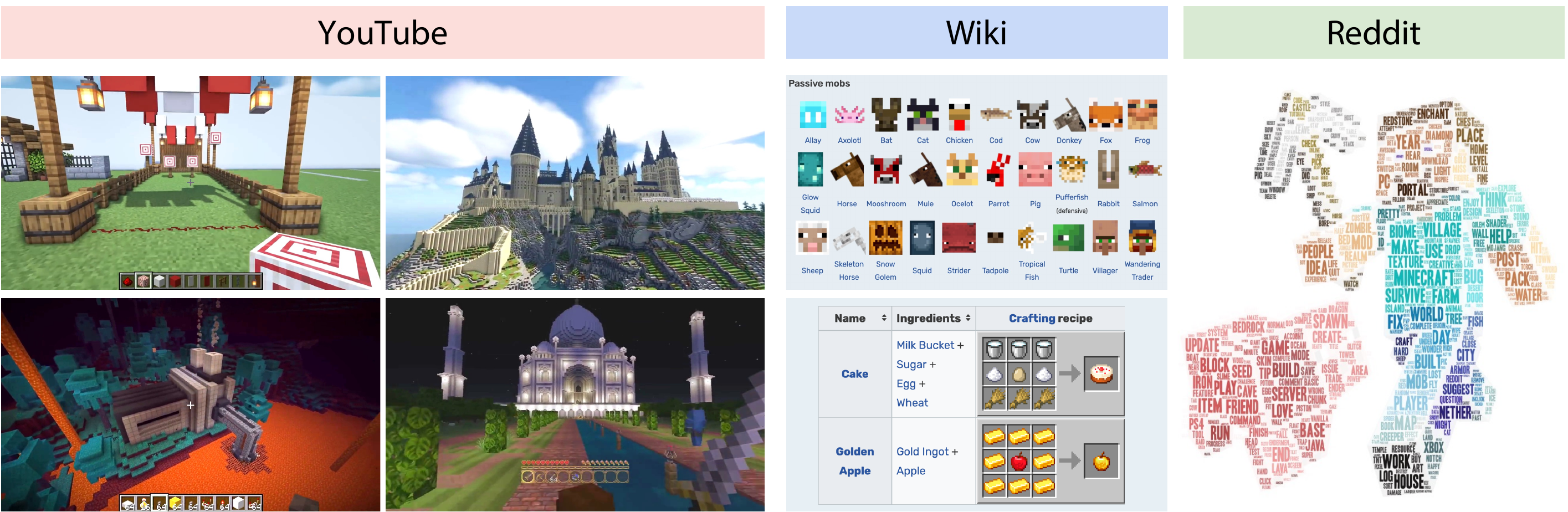}
    \caption{\minedojo's internet-scale, multimodal knowledge base.
    \textbf{Left, YouTube videos:} Minecraft gamers showcase the impressive feats they are able to achieve. Clockwise order: an archery range, Hogwarts castle, Taj Mahal, a Nether homebase.
    \textbf{Middle, Wiki:} Wiki pages contain multimodal knowledge in structured layouts, such as comprehensive catalogs of creatures and recipes for crafting. More examples in Fig.~\ref{supp:fig:wiki_supp} and \ref{supp:fig:wiki_bbox}.  
    \textbf{Right, Reddit:} We create a word cloud from Reddit posts and comment threads. Gamers ask questions, share achievements, and discuss strategies extensively. Sample posts in Fig.~\ref{supp:fig:reddit_supp}. Best viewed zoomed in.
    }
    \label{fig:dataset}
\end{figure}
\section{Internet-scale Knowledge Base}
\label{sec:dataset}

Two commonly used approaches~\cite{silver2017mastering, vinyals2019alphastar, openai2019dota, DBLP:journals/ral/FuchsSKSD21} to train embodied agents include training agents from scratch using RL with well-tuned reward functions for each task, or using a large amount of human-demonstrations to bootstrap agent learning. However, crafting well-tuned reward functions is challenging or infeasible for our task suite (Sec. \ref{sec:benchmark-creative}), and employing expert gamers to provide large amounts of demonstration data would also be costly and infeasible~\cite{vinyals2019alphastar}.

Instead, we turn to the open web as an ever-growing, virtually unlimited source of learning material for embodied agents. The internet provides a vast amount of domain knowledge about Minecraft, which we harvest by extensive web scraping and filtering. We collect 33 years worth of YouTube videos, 6K+ Wiki pages, and millions of Reddit comment threads.  
Instead of hiring a handful of human demonstrators, we capture the collective wisdom of millions of Minecraft gamers around the world. Furthermore, language is a key and pervasive component of our database that takes the form of YouTube transcripts, textual descriptions in Wiki, and Reddit discussions. Language facilitates open-vocabulary understanding, provides grounding for image and video modalities, and unlocks the power of large language models \cite{devlin2018bert,shoeybi2019megatron,brown2020gpt3} for embodied agents. To ensure socially responsible model development, we take special measures to filter out low-quality and toxic contents \cite{bommasani2021foundation,detoxify} from our databases, detailed in the Appendix (Sec.~\ref{supp:sec:dataset}).

\para{YouTube Videos and Transcripts.} 
Minecraft is among the most streamed games on YouTube~\cite{gerblick_2021}. Human players have demonstrated a stunning range of creative activities and sophisticated missions that take hours to complete (examples in Fig. \ref{fig:dataset}). We collect 730K+ narrated Minecraft videos, which add up to 33 years of duration and 2.2B words in English transcripts. 
In comparison, HowTo100M~\cite{miech2019howto100m} is a large-scale human instructional video dataset that includes 15 years of experience in total -- about half of our volume. The time-aligned transcripts enable the agent to ground free-form natural language in video pixels and learn the semantics of diverse activities without laborious human labeling. We operationalize this insight in our pre-trained video-language model (Sec. \ref{sec:method-mineclip-pretraining}). 

\para{Minecraft Wiki.}
The Wiki pages cover almost every aspect of the game mechanics, and supply a rich source of unstructured knowledge in multimodal tables, recipes, illustrations, and step-by-step tutorials. We use Selenium~\cite{selenium} to scrape 6,735 pages that interleave text, images, tables, and diagrams. The pages are highly unstructured and do not share any common schema, as the Wiki is meant for human consumption rather than AI training. To preserve the layout information, we additionally save the screenshots of entire pages and extract 2.2M bounding boxes of the visual elements \Rebuttal{(visualization in Fig.~\ref{supp:fig:wiki_supp} and \mbox{\ref{supp:fig:wiki_bbox}}). We do not use Wiki data in our current experiments. Since the Wiki contains detailed recipes for all crafted objects, they could be provided as input or training data for hierarchical planning methods and policy sketches~\mbox{\cite{andreas2017sketch}}.} Another promising future direction is to apply document understanding models such as LayoutLM~\cite{xu2019layoutlm, xu2020layoutlmv2} and DocFormer~\cite{appalaraju2021docformer} to learn actionable knowledge from these unstructured Wiki data.

\para{Reddit.}
We scrape 340K+ posts along with 6.6M comments under the ``r/Minecraft'' subreddit. These posts ask questions on how to solve certain tasks, showcase cool architectures and achievements in image/video snippets, and discuss general tips and tricks for players of all expertise levels.
\Rebuttal{We do not use Reddit data for training in Sec.~\ref{sec:experiments}, but a potential idea is to finetune large language models \mbox{\cite{devlin2018bert,radford2019gpt2}} on our Reddit corpus to generate instructions and execution plans that are better grounded in the Minecraft domain. Concurrent works~\mbox{\cite{ahn2022saycan, huang2022inner, zeng2022socratic}} have explored similar ideas and showed excellent results on robot learning, which is encouraging for more future research in \mbox{\minedojo}. }
\section{Agent Learning with Large-scale Pre-training}
\begin{figure}
    \centering
    \includegraphics[width=1.01\linewidth]{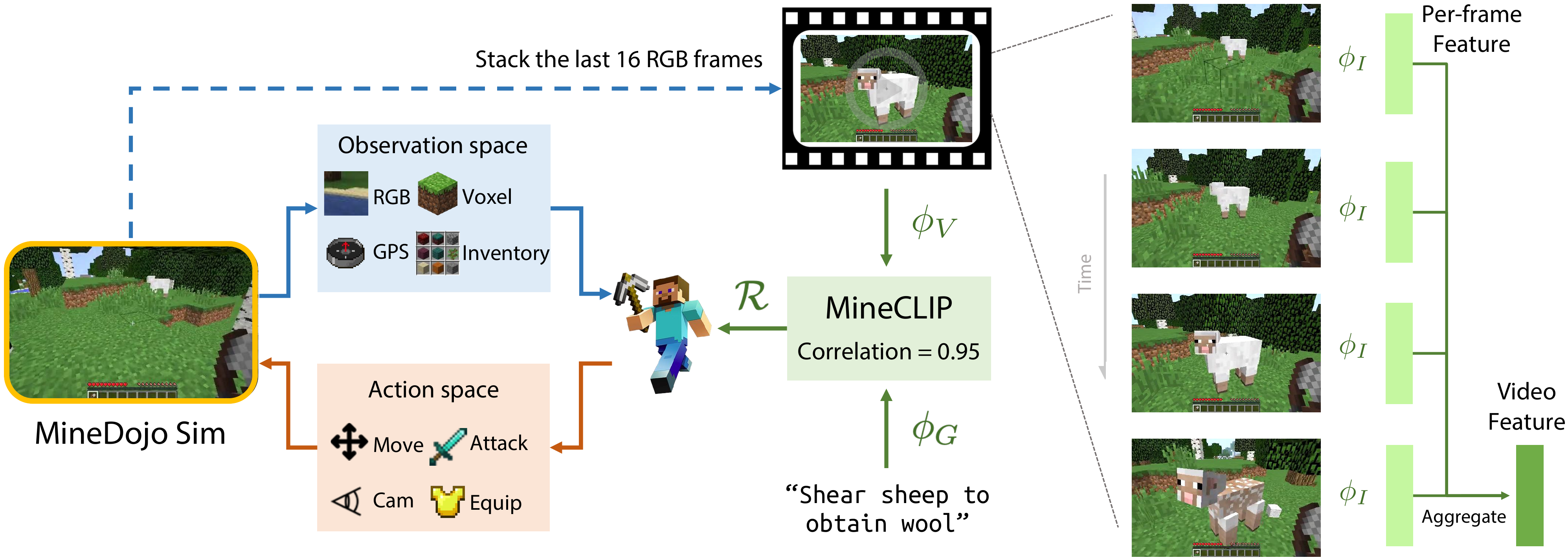}
    \caption{Algorithm design. \mineclip is a contrastive video-language model pre-trained on \minedojo's massive Youtube database. It computes the correlation between an open-vocabulary language goal string and a 16-frame video snippet. The correlation score can be used as a learned dense reward function to train a strong multi-task RL agent.   
    }
    \label{fig:algo_figure}
\end{figure}
\label{sec:method}

One of the grand challenges of embodied AI is to build a single agent that can complete a wide range of open-world tasks.
The \minedojo framework aims to facilitate new techniques towards this goal by providing an open-ended task suite (Sec. \ref{sec:benchmark}) and large-scale internet knowledge base (Sec. \ref{sec:dataset}). 
Here we take an initial step towards this goal by developing a proof of concept that demonstrates how a single language-prompted agent can be trained in \minedojo to complete several complex Minecraft tasks.
To this end, we propose a novel agent learning algorithm that takes advantage of the massive YouTube data offered by \minedojo. We note that this is only one of the numerous possible ways to use \minedojo's internet database --- the Wiki and Reddit corpus also hold great potential to drive new algorithm discoveries for the community in future works.

In this paper, we consider a multi-task reinforcement learning (RL) setting, where an agent is tasked with completing a collection of \minedojo tasks specified by language instructions (Sec. \ref{sec:benchmark}). 
Solving these tasks often requires the agent to interact with the Minecraft world in a prolonged fashion. Agents developed in popular RL benchmarks~\cite{tassa2018deepmind,zhu2020robosuite} often rely on meticulously crafted dense and task-specific reward functions to guide random explorations. However, these rewards are hard or even infeasible to define for our diverse and open-ended tasks in \minedojo.
To address this challenge, our key insight is to learn \textbf{a dense, language-conditioned reward function from in-the-wild YouTube videos and their transcripts}. Therefore, we introduce \textbf{\mineclip}, a contrastive video-language model that learns to correlate video snippets and natural language descriptions (Fig.~\ref{fig:algo_figure}). \mineclip is multi-task by design, as it is trained on open-vocabulary and diverse English transcripts. 

During RL training, \mineclip provides a high-quality reward signal \textit{without} any domain adaptation techniques, despite the domain gap between noisy YouTube videos and clean simulator-rendered frames. 
\mineclip eliminates the need to manually engineer reward functions for each and every \minedojo task.
For Creative tasks that lack a simple success criterion (Sec.~\ref{sec:benchmark-creative}), \mineclip also serves the dual purpose of an \textbf{automatic evaluation metric} that agrees well with human judgement \Rebuttal{on a subset of tasks we investigate} (Sec.~\ref{sec:method-rl}, Table~\ref{table:eval-metric-quality}).  
Because the learned reward model incurs a non-trivial computational overhead, we introduce several techniques to significantly improve RL training efficiency, making \mineclip a practical module for open-ended agent learning in Minecraft (Sec. \ref{sec:method-rl}).

\begin{table*}[t!]
\caption{Our novel \mineclip reward model is able to achieve competitive performance with manually written dense reward function for Programmatic tasks, and significantly outperforms the \openaiclip method across all Creative tasks. \Rebuttal{Entries represent percentage success rates averaged over 3 seeds, each tested for 200 episodes. Success conditions are precise in Programmatic tasks, but estimated by MineCLIP for Creative tasks.}}
\label{table:core-experiments}
\vskip 0.1in
\centering
\resizebox{\textwidth}{!}{%
\begin{tabular}{C{0.1\textwidth}|r|cccccc}

\toprule
Group & Tasks & \textbf{Ours} (Attn) & \textbf{Ours} (Avg) & Manual Reward & Sparse-only & \openaiclip \\ \midrule 

\multirow{4}{*}{\addRoboFig{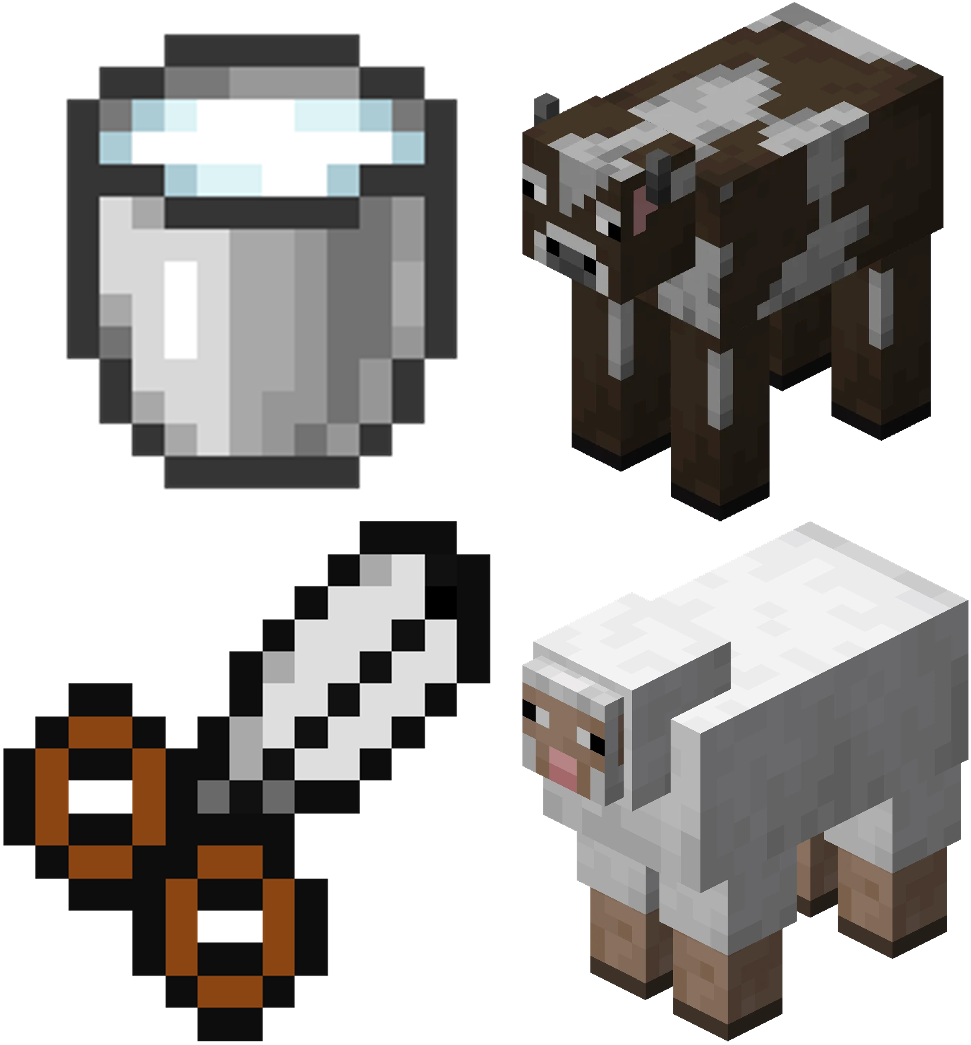}} & Milk Cow & $\bestscore{64.5 \pm 37.1}$ & $6.5 \pm 3.5$ & $62.8 \pm 40.1$ & $0.0 \pm 0.0$ & $0.0 \pm 0.0$ \\
 & Hunt Cow & $\bestscore{83.5 \pm 7.1\hphantom{0}}$ & $0.0 \pm 0.0$ & $48.3 \pm 35.9$ & $0.3 \pm 0.4$ & $0.0 \pm 0.0$ \\
 & Shear Sheep & $12.1 \pm 9.1\hphantom{0}$ & $0.6 \pm 0.2$ & $\bestscore{52.3 \pm 33.2}$ & $0.0 \pm 0.0$ & $0.0 \pm 0.0$ \\
 & Hunt Sheep & $8.1 \pm 4.1$ & $0.0 \pm 0.0$ & $\bestscore{41.9 \pm 33.0}$ & $0.3 \pm 0.4$ & $0.0 \pm 0.0$ \\
\midrule 

\multirow{4}{*}{\addRoboFig{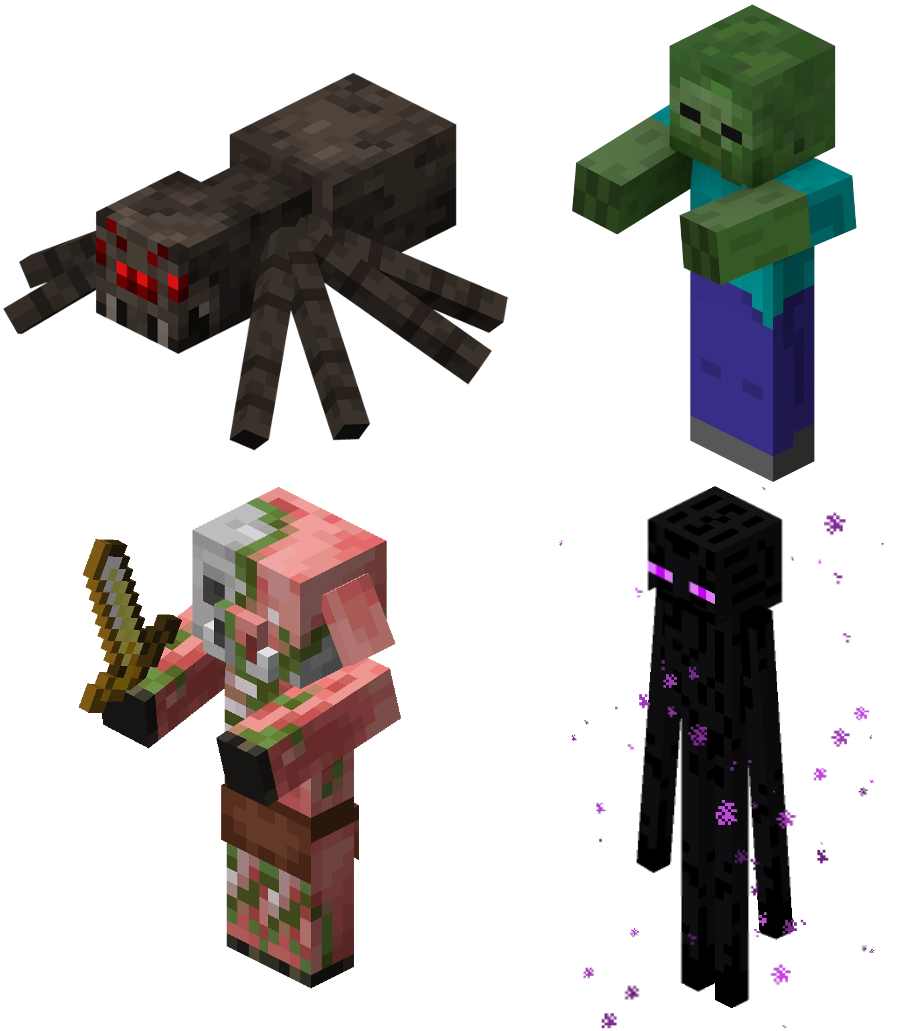}} & Combat Spider &
$80.5 \pm 13.0$ & $60.1 \pm 42.5$ & $\bestscore{87.5 \pm 4.6\hphantom{0}}$ & $47.8 \pm 33.8$ & $0.0 \pm 0.0$ \\
 & Combat Zombie & $47.3 \pm 10.6$ & $\bestscore{72.3 \pm 6.4\hphantom{0}}$ & $49.8 \pm 26.9$ & $\hphantom{0}8.8 \pm 12.4$ & $0.0 \pm 0.0$ \\
 & Combat Pigman & $1.6 \pm 2.3$ & $0.0 \pm 0.0 $ & $\bestscore{13.6 \pm 9.8\hphantom{0}}$ & $0.0 \pm 0.0$ & $0.0 \pm 0.0$ \\
 & Combat Enderman & $0.0 \pm 0.0$ & $0.0 \pm 0.0$ & $0.3 \pm 0.2$ & $0.0 \pm 0.0$ & $0.0 \pm 0.0$\\
\midrule 

\multirow{4}{*}{\addRoboFig{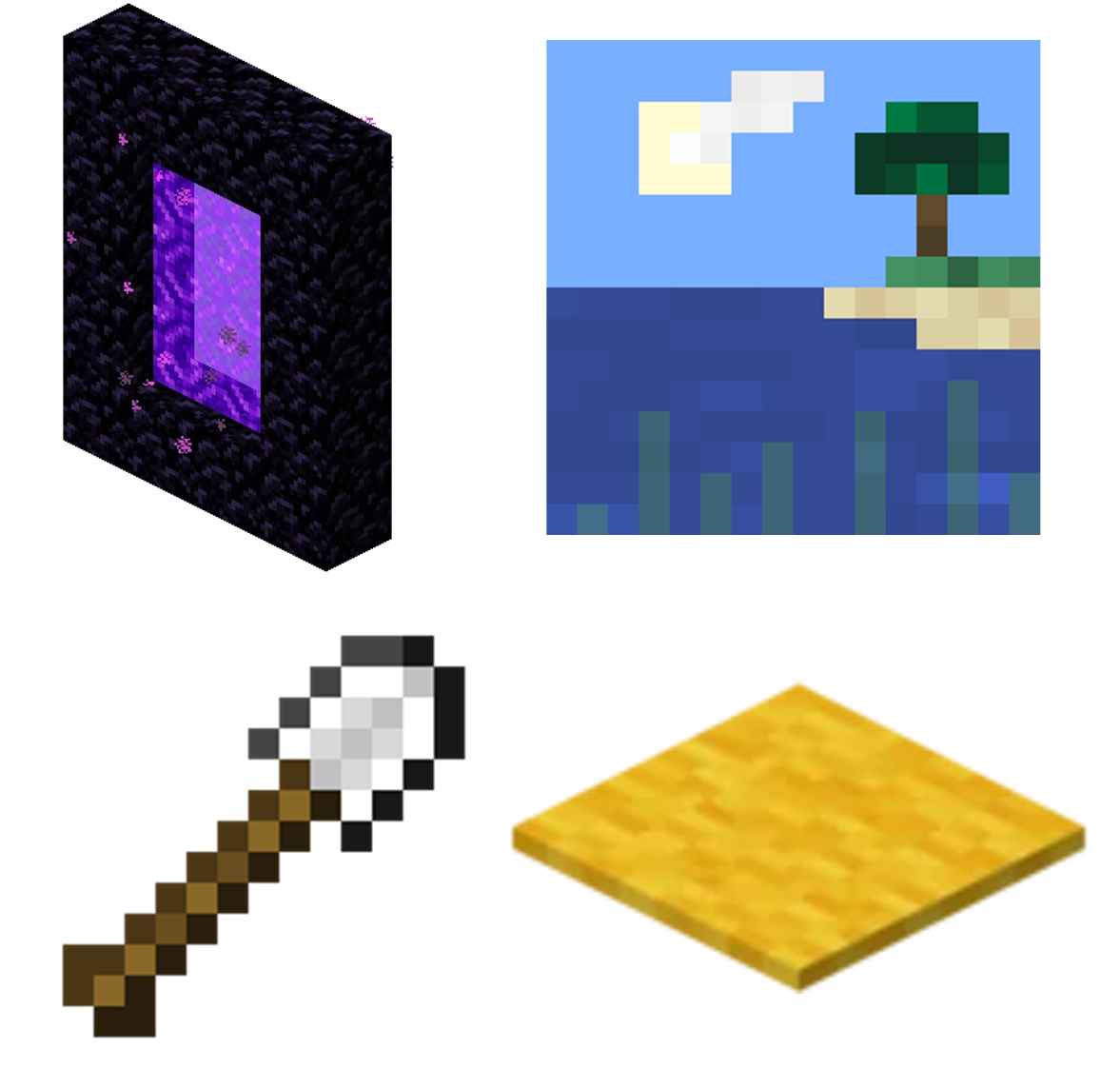}} & Find Nether Portal &
$37.4 \pm 40.8$ & $\bestscore{89.8 \pm 5.7\hphantom{0}}$ & N/A & N/A & $26.3 \pm 32.6$ \\
 & Find Ocean & $33.4 \pm 45.6$ & $\bestscore{54.3 \pm 40.7}$ & N/A & N/A & $\hphantom{0}9.9 \pm 14.1$ \\
 & Dig Hole & $\bestscore{91.6 \pm 5.9\hphantom{0}}$ & $88.1 \pm 13.3$ & N/A & N/A & $0.0 \pm 0.0$ \\
 & Lay Carpet & $97.6 \pm 1.9\hphantom{0}$ & $\bestscore{98.8 \pm 1.0\hphantom{0}}$ & N/A & N/A & $0.0 \pm 0.0$ \\
\bottomrule

\end{tabular}
}
\end{table*}

\subsection{Pre-Training \mineclip on Large-scale Videos}
\label{sec:method-mineclip-pretraining}

Formally, the learned reward function can be defined as $\Phi_\mathcal{R}: (G, V) \rightarrow \mathbb{R}$ that maps a language goal $G$ and a video snippet $V$ to a scalar reward. An ideal $\Phi_\mathcal{R}$ should return a high reward if the behavior depicted in the video faithfully follows the language description, and a low reward otherwise. This can be achieved by optimizing the InfoNCE objective \cite{oord2018representation, DBLP:conf/cvpr/He0WXG20, DBLP:conf/icml/ChenK0H20}, which learns to correlate positive video and text pairs \cite{sun2019learning, DBLP:conf/nips/AlayracRSARFSDZ20, DBLP:conf/cvpr/MiechASLSZ20, akbari2021vatt, xu2021videoclip}. 

Similar to the image-text CLIP model \cite{radford2021clip}, \mineclip is composed of a separate text encoder $\phi_G$ that embeds a language goal and a video encoder $\phi_V$ that embeds a moving window of 16 consecutive frames with $160 \times 256$ resolution (Fig.~\ref{fig:algo_figure}). Our neural architecture has a similar design as CLIP4Clip \cite{luo2021clip4clip}, where $\phi_G$ reuses OpenAI CLIP's pretrained text encoder, and $\phi_V$ is factorized into a frame-wise image encoder $\phi_I$ and a temporal aggregator $\phi_a$ that summarizes the sequence of 16 image features into a single video embedding. Unlike CLIP4Clip, we insert two extra layers of residual CLIP Adapter \cite{gao2021clipadapter} after the aggregator $\phi_a$ to produce a better video feature, and finetune \textit{only} the last two layers of the pretrained $\phi_I$ and $\phi_G$.

From the \minedojo YouTube database, we follow the procedure in VideoCLIP~\cite{xu2021videoclip} to sample 640K pairs of 16-second video snippets and time-aligned English transcripts, after applying a keyword filter. 
We train two \mineclip variants with different types of aggregator $\phi_a$: (1) \mineclipavg does simple average pooling, which is fast but agnostic to the temporal ordering; (2) \mineclipattn encodes the sequence by two transformer layers, which is relatively slower but captures more temporal information, and thus produces a better reward signal in general. 
Details of data preprocessing, architecture, and hyperparameters are listed in the Appendix (Sec.~\ref{supp:sec:mineclip}). 

\subsection{RL with \mineclip Reward}
\label{sec:method-rl}

We train a language-conditioned policy network that takes as input raw pixels and predicts discrete control. The policy is trained with PPO~\cite{schulman2017proximal} on the \mineclip rewards.
In each episode, the agent is prompted with a language goal and takes a sequence of actions to fulfill this goal. When calculating the \mineclip rewards, we concatenate the agent's latest 16 egocentric RGB frames in a temporal window to form a video snippet. \mineclip handles all task prompts \textit{zero-shot} without any further finetuning. In our experiments (Sec. \ref{sec:experiments}), we show that \mineclip provides effective dense rewards out of the box, despite the domain shift between in-the-wild YouTube frames and simulator frames. Besides regular video data augmentation, we do not employ any special domain adaptation methods during pre-training. Our finding is consistent with CLIP's strong zero-shot performances on robustness benchmarks in object recognition \cite{radford2021clip}. 

Compared to hard-coded reward functions in popular benchmarks~\cite{zhu2020robosuite,tassa2018deepmind,fan2021secant}, the \mineclip model has 150M parameters and is thus much more expensive to query. We make several design choices to greatly accelerate RL training with \mineclip in the loop:  

\begin{enumerate}[leftmargin=3em]
    \item The language goal $G$ is fixed for a specific task, so the \textbf{text features $\phi_G$ can be precomputed} to avoid invoking the text encoder repeatedly. 
    \item Our agent's \textbf{RGB encoder reuses the pre-trained weights of $\phi_I$} from \mineclip. We do not finetune $\phi_I$ during RL training, which saves computation and endows the agent with good visual representations from the beginning. 
    \item \mineclip's video encoder $\phi_V$ is factorized into an image encoder $\phi_I$ and a light-weight aggregator $\phi_a$.  This design choice enables \textbf{efficient image feature caching}. Consider two overlapping video sequences of 8 frames, \texttt{V[0:8]} and \texttt{V[1:9]}.
    We can cache the image features of the 7 overlapping frames \texttt{V[1]} to \texttt{V[7]} to maximize compute savings.
    If $\phi_V$ is a monolithic model like S3D \cite{xie2018rethinking} in VideoCLIP \cite{xu2021videoclip}, then the video features from every sliding window must be recomputed, which would incur a much higher cost per time step.
    \item We leverage \textbf{Self-Imitation Learning} \cite{oh2018self} to store the trajectories with high \mineclip reward values in a buffer, and alternate between PPO and self-imitation gradient steps. It further improves sample efficiency as shown in the Appendix (Fig.~\ref{supp:fig:ablation_si}). 
\end{enumerate}

\begin{table}[t!]
\caption{
\Rebuttal{\mbox{\mineclip} agrees well with the ground-truth human judgment on the Creative tasks we consider. Numbers are F1 scores between \mbox{\mineclip}'s binary classification of tasks success and human labels (scaled to the percentage for better readability).}
}
\label{table:eval-metric-quality}
\vskip 0.1in
\centering

\resizebox{0.7\textwidth}{!}{
\begin{tabular}{r|cccccc}
\toprule
 Tasks & Find Nether Portal & Find Ocean & Dig Hole & Lay Carpet  \\ \midrule 

 Ours (Attn) & $98.7$ & $\bestscore{100.0}$ &$99.4$ & $97.4$\\

 Ours (Avg) & $\bestscore{100.0}$ & $\bestscore{100.0}$ & $\bestscore{100.0}$ & $\bestscore{98.4}$\\
 
 \openaiclip & $48.7$ & $98.4$ & $80.6$ & $54.1$\\

\bottomrule
\end{tabular}
}
\end{table}

\section{Experiments}
\label{sec:experiments}

We evaluate our agent-learning approach (Section~\ref{sec:method}) on 8 Programmatic tasks and 4 Creative tasks from the \minedojo benchmarking suite. \Rebuttal{We select these 12 tasks due to the diversity of skills required to solve them (e.g., harvesting, combat, building, navigation) and domain-specific entities (e.g., animals, resources, monsters, terrains, and structures).} We split the tasks into 3 groups and train one multi-task agent for each group: \texttt{Animal-Zoo} (4 Programmatic tasks on hunting or harvesting resource from animals), \texttt{Mob-Combat} (Programmatic, fight 4 types of hostile monsters), and \texttt{Creative} (4 tasks).  

In the experiments, we empirically check the quality of \mineclip against manually written reward functions, and quantify how different variants of our learned model affect the RL performance. Table~\ref{table:core-experiments} presents our main results, and Fig.~\ref{fig:behavior_visualization} visualizes our learned agent behavior in 4 of the considered tasks. 
Policy networks of all methods share the same architecture and are trained by PPO + Self-Imitation (Sec. \ref{sec:method-rl}, training details in the Appendix, Sec.~\ref{supp:sec:rl}). We compare the following methods: 

\begin{itemize}[leftmargin=3em,labelsep=0.7em]
    \item \textbf{Ours (Attn)}: our agent trained with the \mineclipattn reward model. For Programmatic tasks, we also add the final success condition as a binary reward. For Creative tasks, \mineclip is the only source of reward. 
    \item \textbf{Ours (Avg)}: the average-pooling variant of our method.  
    \item \textbf{Manual Reward}: hand-engineered dense reward using ground-truth simulator states. 
    \item \textbf{Sparse-only}: the final binary success as a single sparse reward. Note that neither sparse-only nor manual reward is available for Creative tasks.
    \item \textbf{\openaiclip}: pre-trained OpenAI CLIP model that has \textbf{not} been finetuned on any \minedojo videos. 
\end{itemize}

\para{\mineclip is competitive with manual reward.}
For Programmatic tasks (first 8 rows), RL agents guided by \mineclip achieve competitive performance as those trained by manual reward. In three of the tasks, they even \textit{outperform} the hand-engineered reward functions, which rely on privileged simulator states unavailable to \mineclip. 
\Rebuttal{For a more statistically sound analysis, we conduct the Paired \mbox{Student’s \emph{t}}-test to compare the mean success rate of each task (pairing column 3 ``Ours (Attn)'' and column 5 ``Manual Reward'' in Table \mbox{\ref{table:core-experiments}}). The test yields p-value $0.3991 \gg  0.05$, which indicates that the difference between our method and manual reward is not considered statistically significant, and therefore they are comparable with each other.}
Because all tasks require nontrivial exploration, our approach also dominates the Sparse-only baseline. Note that the original OpenAI CLIP model fails to achieve any success. We hypothesize that the creatures in Minecraft look dramatically different from their real-world counterparts, which causes CLIP to produce \textit{misleading} signals worse than no shaping reward at all. It implies the importance of finetuning on \minedojo's YouTube data. 

\begin{table*}[t!]
\caption{\Rebuttal{\mbox{\mineclip} agents have stronger zero-shot visual generalization ability to unseen terrains, weathers, and lighting. Numbers outside parentheses are percentage success rates averaged over 3 seeds (each tested for 200 episodes), while those inside parentheses are relative performance changes.}}
\label{table:zero-shot-gen-visual}
\vskip 0.1in
\centering
\resizebox{\textwidth}{!}{%
\begin{tabular}{C{0.1\textwidth}|r|cccccc}

\toprule
 & Tasks & \textbf{Ours} (Attn), train & \textbf{Ours} (Attn), unseen test & \openaiclip, train & \openaiclip, unseen test \\ \midrule 

\multirow{4}{*}{\addRoboFig{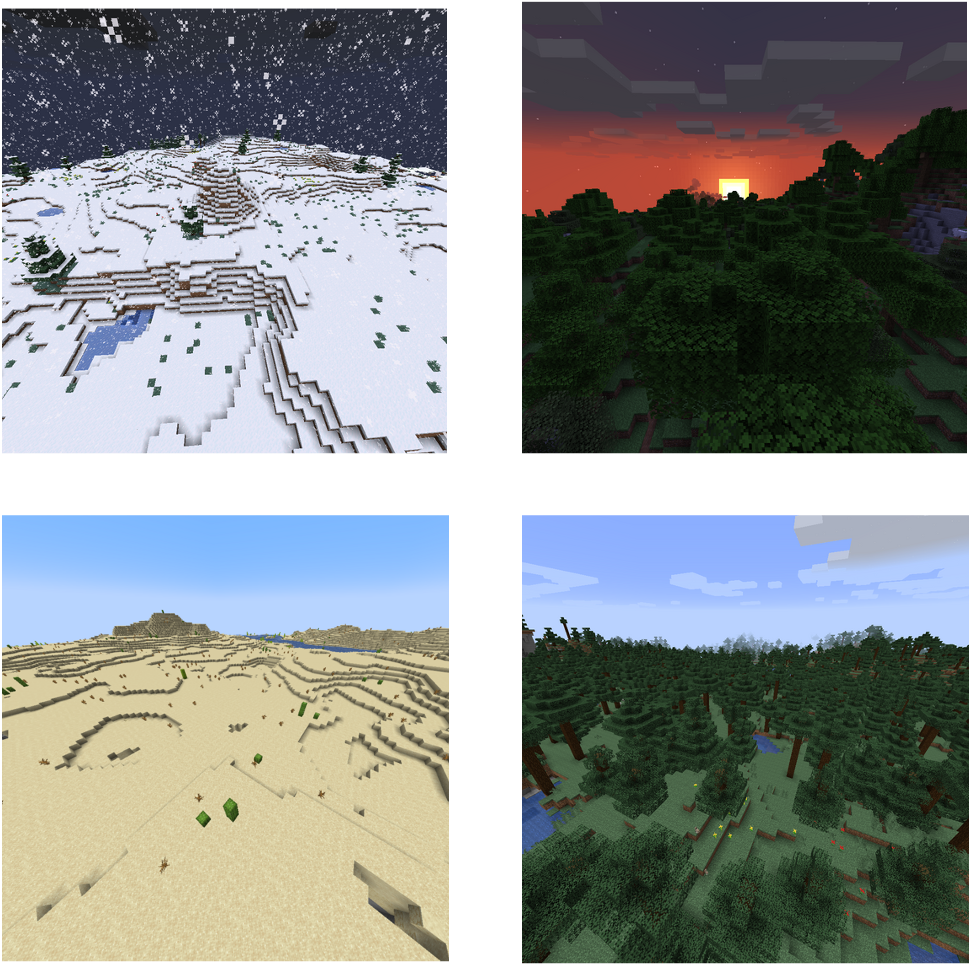}} & Milk Cow & $64.5 \pm 37.1$ & $\bestscore{64.8 \pm 31.3 (+\hphantom{0}0.8\%)}$ & $90.0 \pm 0.4$ & $29.2 \pm 3.7\hphantom{0} (-67.6\%)$ \\
 & Hunt Cow & $83.5 \pm 7.1\hphantom{0}$ & $\bestscore{55.9 \pm 7.2\hphantom{0} (-32.9\%)}$ & $72.7 \pm 3.5$ & $16.7 \pm 1.6\hphantom{0} (-77.0\%)$\\
 & Combat Spider & $80.5 \pm 13.0$ & $\bestscore{62.1 \pm 29.7 (-22.9\%)}$ & $79.5 \pm 2.5$ & $54.2 \pm 9.6\hphantom{0} (-31.8\%)$\\
 & Combat Zombie & $47.3 \pm 10.6$ & $\bestscore{39.9 \pm 25.3 (-15.4\%)}$ & $50.2 \pm 7.5$ & $30.8 \pm 14.4 (-38.6\%)$ \\
\bottomrule

\end{tabular}
}
\end{table*}

\para{\mineclip provides automated evaluation.}
For Creative tasks (last 4 rows), there are no programmatic success criteria available. We convert \mineclip into a binary success classifier by thresholding the reward value it outputs for an episode.  
To test the quality of \mineclip as an automatic evaluation metric, we ask human judges to curate a dataset of 100 successful and 100 failed trajectories for each task. 
We then run both \mineclip variants and \openaiclip on the dataset and report the binary F1 score of their judgement against human ground-truth in Table \ref{table:eval-metric-quality}. 
The results demonstrate that both \mineclipattn and \mineclipavg attain a very high degree of agreement with human evaluation results \Rebuttal{on this subset of the Creative task suite that we investigate}. \openaiclip baseline also achieves nontrivial agreement on Find Ocean and Dig Hole tasks, likely because real-world oceans and holes have similar texture. We use the \texttt{attn} variant as an automated success criterion to score the 4 Creative task results in Table \ref{table:core-experiments}. Our proposed method consistently learns better than \openaiclip-guided agents. It shows that \mineclip is an effective approach to solving open-ended tasks when no straightforward reward signal is available. \Rebuttal{We provide further analysis beyond these 4 tasks in the 
Appendix (Sec.~\ref{supp:sec:mineclip_complex_creative})}.

\para{\mineclip shows good zero-shot generalization to significant visual distribution shift.}
\Rebuttal{We evaluate the learned policy without finetuning on a combination of unseen weather, lighting conditions, and terrains --- 27 scenarios in total. For the baseline, we train agents with the original \mbox{\openaiclip} image encoder (not trained on our YouTube videos) by imitation learning. The robustness against visual shift can be quantitatively measured by the relative performance degradation on novel test scenarios for each task. Table~\mbox{\ref{table:zero-shot-gen-visual}} shows that while all methods incur performance drops, agents with \mbox{\mineclip} encoder is more robust to visual changes than the baseline across all tasks. Pre-training on diverse in-the-wild YouTube videos is important to boosting zero-shot visual generalization capability, a finding consistent with literature \mbox{\cite{radford2021clip, nair2022r3m}}.  
}

\para{Learning a Single Agent for All 12 Tasks}
We have also trained a single agent for all 12 tasks. To reduce the computational burden without loss of generality, the agent is trained by self-imitating from successful trajectories generated from the self-imitation learning pipeline (Section \ref{sec:appendix_rl_training}). We summarize the result in Table \ref{table:supp-single-agent-12-tasks}. Similar to our main experiments, all numbers represent percentage success rates averaged over 3 training seeds, each tested for 200 episodes. 
Compared to the original agents, the new 12-multitask agent sees a performance boost in 6 tasks, degradation in 4 tasks, and roughly the same success rates in 2 tasks. This result suggests that there are both positive and negative task transfers happening. To improve the multi-task performance, more advanced algorithms \cite{yu2020gradient,wu2020understanding} can be employed to mitigate the negative transfer effects.

We also perform Paired Student’s \textit{t}-test to statistically compare the performance of the 12-multitask agent and those separately trained on each task group. We obtain a p-value of $0.3720 \gg 0.05$, which suggests that the difference between the two training settings is not statistically significant.

\begin{table*}[t!]
\caption{\Rebuttal{We train a single multi-task agent for all 12 tasks. All numbers represent percentage success rates averaged over 3 seeds, each tested for 200 episodes.}}
\label{table:supp-single-agent-12-tasks}
\vskip 0.1in
\centering
\resizebox{\textwidth}{!}{%
\begin{tabular}{C{0.1\textwidth}|r|cccccc}

\toprule
Group & Tasks & Single Agent on All Tasks & Original & Performance Change \\ \midrule 

\multirow{4}{*}{\addRoboFig{task1}} & Milk Cow & $\bestscore{91.5 \pm 0.7}$ & $64.5 \pm 37.1$ & $\uparrow$\\
 & Hunt Cow & $46.8 \pm 3.7$ & $\bestscore{83.5 \pm 7.1\hphantom{0}}$ & $\downarrow$\\
 & Shear Sheep & $\bestscore{73.5 \pm 0.8}$ & $12.1 \pm 9.1\hphantom{0}$ & $\uparrow$\\
 & Hunt Sheep & $\bestscore{27.0 \pm 1.0}$ & $8.1 \pm 4.1$ & $\uparrow$\\
\midrule 

\multirow{4}{*}{\addRoboFig{task2}} & Combat Spider & $72.1 \pm 1.3$ &
$\bestscore{80.5 \pm 13.0}$ & $\downarrow$\\
 & Combat Zombie & $27.1 \pm 2.7$ & $\bestscore{47.3 \pm 10.6}$ & $\downarrow$\\
 & Combat Pigman & $\bestscore{6.5 \pm 1.2}$ & $1.6 \pm 2.3$ & $\uparrow$\\
 & Combat Enderman & $0.0 \pm 0.0$ & $0.0 \pm 0.0$ & $=$\\
\midrule 

\multirow{4}{*}{\addRoboFig{task3}} & Find Nether Portal & $\bestscore{99.1 \pm 0.4}$ &
$37.4 \pm 40.8$ & $\uparrow$\\
 & Find Ocean & $\bestscore{95.1 \pm 1.5}$ & $33.4 \pm 45.6$ & $\uparrow$\\
 & Dig Hole & $85.8 \pm 1.2$ & $\bestscore{91.6 \pm 5.9\hphantom{0}}$ & $\downarrow$\\
 & Lay Carpet & $96.5 \pm 0.8$ & $\bestscore{97.6 \pm 1.9}\hphantom{0}$ & $=$\\
\bottomrule

\end{tabular}
}
\vspace{0.1in} 
\end{table*}

\begin{table*}[t!]
\caption{\Rebuttal{We test the open-vocabulary generalization ability to two unseen tasks. All numbers represent percentage success rates averaged over 3 seeds, each tested for 200 episodes.}}
\label{table:supp-hunt-pig-and-spider-string}
\vskip 0.1in
\centering
\resizebox{\textwidth}{!}{%
\begin{tabular}{C{0.1\textwidth}|r|cccccc}

\toprule
 & Tasks & Ours (zero-shot) & Ours (after RL finetune) & Baseline (RL from scratch) \\ \midrule 

\multirow{2}{*}{\addRoboFig{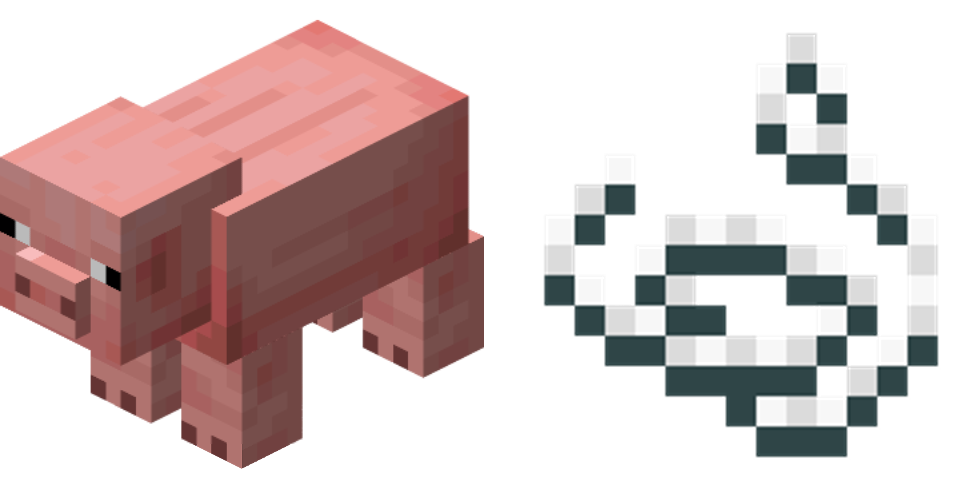}} & Hunt Pig & $1.3 \pm 0.6$ & $\bestscore{46.0 \pm 15.3}$ & $0.0 \pm 0.0$\\
 & Harvest Spider String & $1.6 \pm 1.3$ & $\bestscore{36.5 \pm 16.9}$ & $12.5 \pm 12.7$\\

\bottomrule

\end{tabular}
}
\end{table*}

\para{Generalize to Novel Tasks}
To test the ability to generalize to new open-vocabulary commands, we evaluate the agent on two novel tasks: ``harvest spider string'' and ``hunt pig''. Table \ref{table:supp-hunt-pig-and-spider-string} shows that the agent struggles in the zero-shot setting because it has not interacted with pigs or spider strings during training, and thus does not know how to interact with them effectively. However, the performance improves substantially by finetuning with the \mineclip reward. Here the baseline methods are trained from scratch using RL with the \mineclip encoders and reward. Therefore, the only difference is whether the policy has been pre-trained on the 12 tasks or not. Given the same environment sampling budget (only around 5\% of total samples), it significantly outperforms baselines. It suggests that the multitask agent has learned transferable knowledge on hunting and resource collection, which enables it to quickly adapt to novel tasks.

\section{Related work}
\label{sec:related}
\para{Open-ended Environments for Decision-making Agents.} 
There are many environments developed with the goal of open-ended agent learning. Prior works include maze-style worlds \cite{team2021openended,wang2019paired, juliani2019obstacletower}, purely text-based game~\cite{heinrich2020nethack}, grid worlds ~\cite{boisvert2019babyai,cao2020babyai}, browser/GUI-based environments ~\cite{shi2017worldofbits,toyama2021androidenv}, and indoor simulators for robotics ~\cite{deepmind2020playroom,shen2020igibson,srivastava2021behavior,fan2021secant,shridhar2020alfred,savva2019habitat,puig2018virtualhome}. 
Minecraft offers an exciting alternative for open-ended agent learning. It is a 3D visual world with procedurally generated landscapes and extremely flexible game mechanics that support an enormous variety of activities.
Prior methods in open-ended agent learning \Rebuttal{\mbox{\cite{  ecoffet2019goexplore, huizinga2022evolving, wang2020enhanced,  kanitscheider2021multitask, dennis2020paired}} do not make use of external knowledge, but our approach leverages internet-scale database to learn open-vocabulary reward models, thanks to Minecraft's abundance of gameplay data online}.  

\vspace{-0.05in}
\para{Minecraft for AI Research.} The Malmo platform \cite{johnson16malmo} is the first comprehensive release of a Gym-style agent API \cite{brockman2016openai} for Minecraft. Based on Malmo, MineRL \cite{guss2019minerl} provides a codebase and human play trajectories for the annual Diamond Challenge at NeurIPS \cite{guss2019minerl2,guss2021minerl,kanervisto2022minerl}. \minedojo's simulator builds upon the pioneering work of MineRL, but greatly expands the API and benchmarking task suite. Other Minecraft benchmarks exist with different focuses. For example, CraftAssist~\cite{gray2019craftassist} and IGLU~\cite{kiseleva2021neurips} study agents with interactive dialogues.  BASALT~\cite{shah2021basalt} applies human evaluation to 4 open-ended tasks. EvoCraft~\cite{grbic21evocraft} is designed for structure building, and Crafter~\cite{hafner2021crafter} optimizes for fast experimentation. Unlike prior works, \minedojo's core mission is to facilitate the development of generally capable embodied agents using internet-scale knowledge. \Rebuttal{We include a feature comparison table of different Minecraft platforms for AI research in Table~\ref{supp:table:framework_comparison}}.

\vspace{-0.05in}
\para{Internet-scale Multimodal Knowledge Bases.} 
Big dataset such as Common Crawl~\cite{commoncrawl}, the Pile \cite{gao2020pile}, LAION \cite{schuhmann2021laion}, YouTube-8M \cite{haija2016youtube8m} and HowTo100M~\cite{miech2019howto100m} have been fueling the success of large pre-trained language models~\cite{devlin2018bert,radford2019gpt2, brown2020gpt3} and multimodal models~\cite{sun2019learning, DBLP:conf/nips/AlayracRSARFSDZ20, DBLP:conf/cvpr/MiechASLSZ20, zhu2020actbert, DBLP:conf/aaai/AmraniBRB21, akbari2021vatt, xu2021videoclip}. While generally useful for learning representations, these datasets are not specifically targeted at embodied agents. To provide agent-centric training data, RoboNet~\cite{dasari19robonet} collects video frames from 7 robot platforms, and Ego4D~\cite{grauman2021ego4d} recruits volunteers to record egocentric videos of household activities. In comparison, \minedojo's knowledge base is constructed without human curation efforts, much larger in volume, more diverse in data modalities, and comprehensively covers all aspects of the Minecraft environment. 

\vspace{-0.05in}
\para{Embodied Agents with Large-scale Pre-training.} 
Inspired by the success in NLP, embodied agent research\Rebuttal{~\mbox{\cite{duan2022survey, batra2020rearrangement, ravichandar2020recent, collins2021review}}} has seen a surge in adoption of the large-scale pre-training paradigm. The recent advances can be roughly divided into 4 categories. 
1) \textbf{Novel agent architecture}: Decision Transformer \cite{ranzato2021decisiontransformer,janner2021onebigsequence,zheng2022onlinedt} applies the powerful self-attention models to sequential decision making. \Rebuttal{GATO~\mbox{\cite{reed2022generalist}} and Unified-IO~\mbox{\cite{lu2022unifiedio}} learn a single model to solve various decision-making tasks with different control interfaces.} VIMA~\cite{jiang2022vima} unifies a wide range of robot manipulation tasks with multimodal prompting. 
2)  \textbf{Pre-training for better representations}: R3M~\cite{nair2022r3m} trains a general-purpose visual encoder for robot perception on Ego4D videos~\cite{grauman2021ego4d}. CLIPort~\cite{shridhar2021cliport} leverages the pre-trained CLIP model \cite{radford2021clip} to enable free-form language instructions for robot manipulation. 
3) \textbf{Pre-training for better policies}: AlphaStar \cite{vinyals2019alphastar} achieves champion-level performance on StarCraft by imitating from numerous human demos. SayCan \cite{ahn2022saycan} leverages large language models (LMs) to ground value functions in the physical world. \cite{li2022pretrained} and \cite{reid2022wikipedia} directly reuse pre-trained LMs as policy backbone. \Rebuttal{VPT~\mbox{\cite{openai2022vpt}} is a concurrent work that learns an inverse dynamics model from human contractors to pseudo-label YouTube videos for behavior cloning. VPT is complementary to our approach, and can be finetuned to solve language-conditioned open-ended tasks with our learned reward model. } 
4) \textbf{Data-driven reward functions}: Concept2Robot~\cite{shao2021concept2robot} and DVD~\cite{nair2021dvd} learn a binary classifier to score behaviors from in-the-wild videos~\cite{goyal2017something}. LOReL~\cite{nair2021lorel} crowd-sources humans labels to train language-conditioned reward function for offline RL. AVID \cite{smith2019avid} and XIRL \cite{zakka2021xirl} extract reward signals via cycle consistency. 
\minedojo's task benchmark and internet knowledge base are generally useful for developing new algorithms in all the above categories. In Sec. \ref{sec:method}, we also propose an open-vocabulary, multi-task reward model using \minedojo YouTube videos.

\section{Conclusion}
\label{sec:conclusion}

In this work, we introduce the \minedojo framework for developing generally capable embodied agents. \minedojo features a benchmarking suite of thousands of Programmatic and Creative tasks, and an internet-scale multimodal knowledge base of videos, wiki, and forum discussions. As an example of the novel research possibilities enabled by \minedojo, we propose \mineclip as an effective language-conditioned reward function trained with in-the-wild YouTube videos. \mineclip achieves strong performance empirically and agrees well with human evaluation results, making it a good automatic metric for Creative tasks. We look forward to seeing how \minedojo empowers the community to make progress on the important challenge of open-ended agent learning.

\section{Acknowledgement}
\label{sec:acknowledgement}

We are extremely grateful to Anssi Kanervisto, Shikun Liu, Zhiding Yu, Chaowei Xiao, Weili Nie, Jean Kossaifi, Jonathan Raiman, Neel Kant, Saad Godil,  Jaakko Haapasalo, Bryan Catanzaro, John Spitzer, Zhiyuan ``Jerry'' Lin, Yingqi Zheng, Chen Tessler, Dieter Fox, Oli Wright, Jeff Clune, Jack Parker-Holder, and many other colleagues and friends for their helpful feedback and insightful discussions. We also thank the anonymous reviewers for offering us highly constructive advice and kind encouragement during the review and rebuttal period. NVIDIA provides the necessary computing resource and infrastructure for this project. Guanzhi Wang is supported by the Kortschak fellowship in Computing and Mathematical Sciences at Caltech. 
\bibliography{ms}
\bibliographystyle{plainnat}

\newpage
\appendix
\newcommand{\direct}[0]{\textsc{Direct}\xspace}
\newcommand{\rdelta}[0]{\textsc{Delta}\xspace}
\renewcommand{\thefigure}{A.\arabic{figure}}
\setcounter{figure}{0}
\renewcommand{\thetable}{A.\arabic{table}}
\setcounter{table}{0}

\section{Minecraft Framework Comparison}
\begin{table}[!h]
\caption{\Rebuttal{Comparison table of different Minecraft platforms for AI research.}}
\label{supp:table:framework_comparison}
\vspace{0.1in}
\resizebox{1.0\textwidth}{!}{%
\begin{tabular}{p{0.1\textwidth}p{0.3\textwidth}p{0.05\textwidth}p{0.05\textwidth}p{0.02\textwidth}p{0.3\textwidth}p{0.3\textwidth}}
\toprule
\multicolumn{1}{l}{\multirow{2}{*}{Environment}} & \multicolumn{2}{c}{Simulator} & \multicolumn{2}{c}{Task Suite} & \multicolumn{2}{c}{Knowledge Base} \\ 
\cmidrule(lr){2-3}\cmidrule(lr){4-5}\cmidrule(lr){6-7}
\multicolumn{1}{c}{} & \multicolumn{1}{c}{Features} & \multicolumn{1}{c}{\makecell[cb]{Real \\ Minecraft}} & \multicolumn{1}{c}{\makecell[cb]{Number \\ of Tasks}} & \multicolumn{1}{c}{\makecell[cb]{Language-\\grounded}}& \multicolumn{1}{c}{Features} & \multicolumn{1}{c}{Data Scale}\\
\midrule
\minedojo & Unified observation and action space; \newline unlocks all three types of world (the Overworld, the Nether, and the End) & \multicolumn{1}{c}{\checkmark} & $3,000+$ & \multicolumn{1}{c}{\checkmark} & Automatically scraped from the Internet;\newline multimodal data (videos, images, texts, tables and diagrams) & 740K YouTube videos;\newline 7K Wiki pages;\newline 350K Reddit posts\\
\midrule
MineRL v0.4~\mbox{\cite{guss2019minerl}} & Built on top of Malmo;\newline actively maintained & \multicolumn{1}{c}{\checkmark} & 11 & & Annotated state-action pairs of human demonstrations & 60M frames of recorded human player data\\
\midrule
MineRL v1.0 (VPT)~\mbox{\cite{openai2022vpt}} & Mouse and keyboard control & \multicolumn{1}{c}{\checkmark} & 5 & & Labeled contractor data;\newline unlabeled videos scraped from the Internet & 2K hours of contractor data; \newline 270K hours of unlabeled videos\\
\midrule
MarLÖ~\mbox{\cite{perez-liebana2019multiagent}} & Cooperative and competitive multiagent tasks;\newline parameterizable environments & \multicolumn{1}{c}{\checkmark} & 14 & & & \\
\midrule
Malmo~\mbox{\cite{johnson16malmo}} & First comprehensive release of a Gym-style agent API for Minecraft & \multicolumn{1}{c}{\checkmark} & N/A & & & \\
\midrule
CraftAssist~\mbox{\cite{gray2019craftassist}} & Bot assistant;\newline  dialogue interactions & \multicolumn{1}{c}{\checkmark} & N/A & \multicolumn{1}{c}{\checkmark} & Interactive dialogues;\newline crowd-sourced house building dataset & 800K dialogue-action dictionary pairs;\newline 2.6K houses with atomic building actions\\
\midrule
IGLU~\mbox{\cite{kiseleva2021neurips}} & Interactive dialogues with humans;\newline aimed at building structures described by natural language & & 157 & \multicolumn{1}{c}{\checkmark} & & \\
\midrule
EvoCraft~\mbox{\cite{grbic21evocraft}} & Aimed at generating creative artifacts;\newline allows for direction manipulation of blocks & & N/A & & & \\
\midrule
Crafter~\mbox{\cite{hafner2021crafter}} & 2D clone of Minecraft;\newline fast experimentation & & 22 & & Human experts dataset & 100 episodes\\
\bottomrule
\end{tabular}
}
\end{table}
\vspace{0.1in}

\section{\minedojo Simulator}
\label{supp:sec:sim}

We design unified observation and action spaces across all tasks to facilitate the development of multi-tasking and continually learning agents that can adapt to novel tasks and scenarios. The codebase is open sourced at \href{https://github.com/MineDojo/MineDojo}{github.com/MineDojo/MineDojo}. 

\subsection{Observation Space}
\label{supp:sec:sim_obs_space}

Our observation space contains multiple modalities. The agent perceives the world mainly through the RGB screen. To provide the same information as human players receive, we also supplement the agent with observations about its inventory, location, health, surrounding blocks, etc. The full observation space is shown below. We refer readers to see our code documentation for technical details such as data type for each observable item.

We also support a LIDAR sensor that returns the groundtruth type of the blocks that the agent sees, however this is considered privileged information and does not go into the benchmarking specification. However, it is still useful for hand engineering the dense reward function, which we use in our experiments (Sec.~\ref{sec:experiments}). Amounts and directions of LIDAR rays can be arbitrarily configured at the cost of a lower simulation throughput.

\vspace{0.2in}
\resizebox{\textwidth}{!}{%

\begin{tabular}{r|l|l}
\toprule
\textbf{Modality}       & \textbf{Shape} & \textbf{Description}                                                  \\ \midrule
RGB                 & \texttt{(3, H, W)}      & Ego-centric RGB frames.                                               \\
Equipment           & \texttt{(6,)}           & Names, quantities, variants, and durabilities of equipped items.      \\
Inventory           & \texttt{(36,)}          & Names, quantities, variants, and durabilities of inventory items.     \\
Voxel               & \texttt{(3, 3, 3)}      & Names, variants, and properties of $3 \times 3 \times 3$ surrounding blocks.                \\
Life statistics     & \texttt{(1,)}           & Agent's health, oxygen, food saturation, etc.                         \\
GPS                 & \texttt{(3,)}          & GPS location of the agent.                                            \\
Compass             & \texttt{(2,)}           & Yaw and pitch of the agent.                                           \\
Nearby tools        & \texttt{(2,)}           & Indicate if crafting table and furnace are nearby the agent.          \\
Damage source       & \texttt{(1,)}          & Information about the damage on the agent.                            \\
Lidar               & \texttt{(Num rays,)}     & Ground-truth lidar observation.                                       \\ \bottomrule
\end{tabular}

}

\subsection{Action Space}
\label{supp:sec:sim_action_space}

We design a compound action space. At each step the agent chooses one movement action (forward, backward, camera actions, etc.) and one optional functional action as listed in the table below. Some functional actions such as \texttt{craft} take one argument, while others like \texttt{attack} does not take any argument. This compound action space can be modelled in an autoregressive manner~\cite{vinyals2019alphastar}. We refer readers to our code documentation for example usages of our action space.

\begin{center}
    
\begin{tabular}{@{}r|l|l@{}}
\toprule
\textbf{Name} & \textbf{Description}     & \textbf{Argument}                 \\ \midrule
\texttt{no\_op}         & Do nothing.    & $\varnothing$                      \\
\texttt{use}           & Use the item held in the main hand.  &   $\varnothing$                \\
\texttt{drop}          & Drop the item held in the main hand.                & $\varnothing$                      \\
\texttt{attack}        & Attack with barehand or tool held in the main hand. & $\varnothing$                      \\
\texttt{craft}         & Execute a crafting recipe to obtain new items.      & Index of recipe    \\
\texttt{equip}         & Equip an inventory item.                            & Slot index of the item \\
\texttt{place}         & Place an inventory item on the ground.              & Slot index of the item \\
\texttt{destroy}       & Destroy an inventory item.                          & Slot index of the item \\ \bottomrule
\end{tabular}
\end{center}

\subsection{Customizing the Environment}
Environments in \mineclip simulator can be easily and flexibly customized. Through our simulator API, users can control terrain, weather, day-night condition (different lighting), the spawn rate and range of specified entities and materials, etc. We support a wide range of terrains, such as desert, jungle, taiga, and iced plain, and special in-game structures, such as ocean monument, desert temple, and End city.
Please visit our website for video demonstrations.

\section{\minedojo Task Suite}
\label{supp:sec:task_suite}

In this section, we explain how we collect the Programmatic (Sec.~\ref{sec:benchmark-programmatic}) and Creative tasks (Sec.~\ref{sec:benchmark-creative}). 

\definecolor{codegreen}{HTML}{76b900}
\definecolor{codegray}{rgb}{0.5,0.5,0.5}
\definecolor{codepurple}{rgb}{0.58,0,0.82}
\definecolor{backcolour}{rgb}{0.95,0.95,0.92}

\lstdefinestyle{mystyle}{
    keywords={category, prompt, survival_sword_food, harvest_wool_with_shears_and_sheep, techtree_from_barehand_to_wooden_sword, combat_zombie_pigman_nether_diamond_armors_diamond_sword_shield},
    backgroundcolor=\color{backcolour},   
    commentstyle=\color{codegreen},
    keywordstyle=\color{magenta},
    numberstyle=\tiny\color{codegray},
    stringstyle=\color{codepurple},
    basicstyle=\ttfamily\footnotesize,
    breakatwhitespace=false,         
    breaklines=true,                 
    captionpos=b,                    
    keepspaces=true,                 
    numbers=left,                    
    numbersep=5pt,                  
    showspaces=false,                
    showstringspaces=false,
    showtabs=false,                  
    tabsize=2
}

\lstset{style=mystyle}

\begin{figure}[t!]

\begin{lstlisting}
survival_sword_food:
  category: survival
  prompt: survive as long as possible given a sword and some food
  
harvest_wool_with_shears_and_sheep:
  category: harvest
  prompt: harvest wool from a sheep with shears and a sheep nearby

techtree_from_barehand_to_wooden_sword:
  category: tech-tree
  prompt: find material and craft a wooden sword
  
combat_zombie_pigman_nether_diamond_armors_diamond_sword_shield:
  category: combat
  prompt: combat a zombie pigman in nether with a diamond sword,
    shield, and a full suite of diamond armors
\end{lstlisting}

\caption{Example specifications.}
\label{supp:fig:prompt_example}
\end{figure}
\vspace{0.1in}

\subsection{Programmatic Tasks}
Programmatic tasks are constructed by filling manually written templates for four categories of tasks, namely ``Survival'', ``Harvest'', ``Tech Tree'', and ``Combat''. The task specifications are included in our codebase. Please refer to Fig. \ref{supp:fig:prompt_example} for a few samples.
We briefly explain each task category: 

\para{Survival.}
This task group tests the ability to stay alive in the game. It is nontrivial to survive in Minecraft, because the agent grows hungry as time passes and the health bar drops gradually. Hostile mobs like zombie and skeleton spawn at night, which are very dangerous if the agent does not have the appropriate armor to protect itself or weapons to fight back. We provide two tasks with different levels of difficulty for Survival. One is to start from scratch without any assistance. The other is to start with initial weapons and food. 

\para{Harvest.}
This task group tests the agent's ability to collect useful resources such as minerals (iron, diamond, obsidian), food (beef, pumpkin, carrots, milk), and other useful items (wool, oak wood, coal). We construct these tasks by enumerating the Cartesian product between target items to collect, initial inventory, and world conditions (terrain, weather, lighting, etc.) so that they cover a spectrum of difficulty. 
For instance, if the task is to harvest wool, then it is relatively easy if the agent has a shear in its initial inventory with a sheep nearby, but more difficult if the agent has to craft the shear from raw material and explore extensively to find a sheep.
We filter out combinations that are impossible (such as farming certain plants in the desert) from the Cartesian product.

\para{Tech Tree.}
Minecraft includes several levels of tools and armors with different properties and difficulties to unlock. To progress to a higher level of tools and armors, the agent needs to develop systematic and compositional skills to navigate the technology tree (e.g. wood $\rightarrow$ stone $\rightarrow$ iron $\rightarrow$ diamond). 
In this task group, the agent is asked to craft and use a hierarchy of tools starting from a less advanced level. For example, some task asks the agent to craft a wooden sword from bare hand. Another task may ask the agent to craft a gold helmet. An agent that can successfully complete these tasks should have the ability to transfer similar exploration strategies to different tech levels.

\para{Combat.}
We test agent's reflex and martial skills to fight against various monsters and creatures. Similar to how we develop the Harvest task group, we generate these tasks by enumerating the Cartesian product between the target entity to combat with, initial inventory, and world conditions to cover a spectrum of difficulty.

\subsection{Creative Tasks}
\label{supp:sec:creative_tasks}

We construct Creative tasks using three approaches: 1) manual brainstorming, 2) mining from YouTube tutorial videos, and 3) generate by querying GPT-3 API. We elaborate the second and third approaches below.

\para{Task Mining from YouTube Tutorial Videos.}
\begin{figure}[t!]
    \centering
    \includegraphics[width=1.0\linewidth]{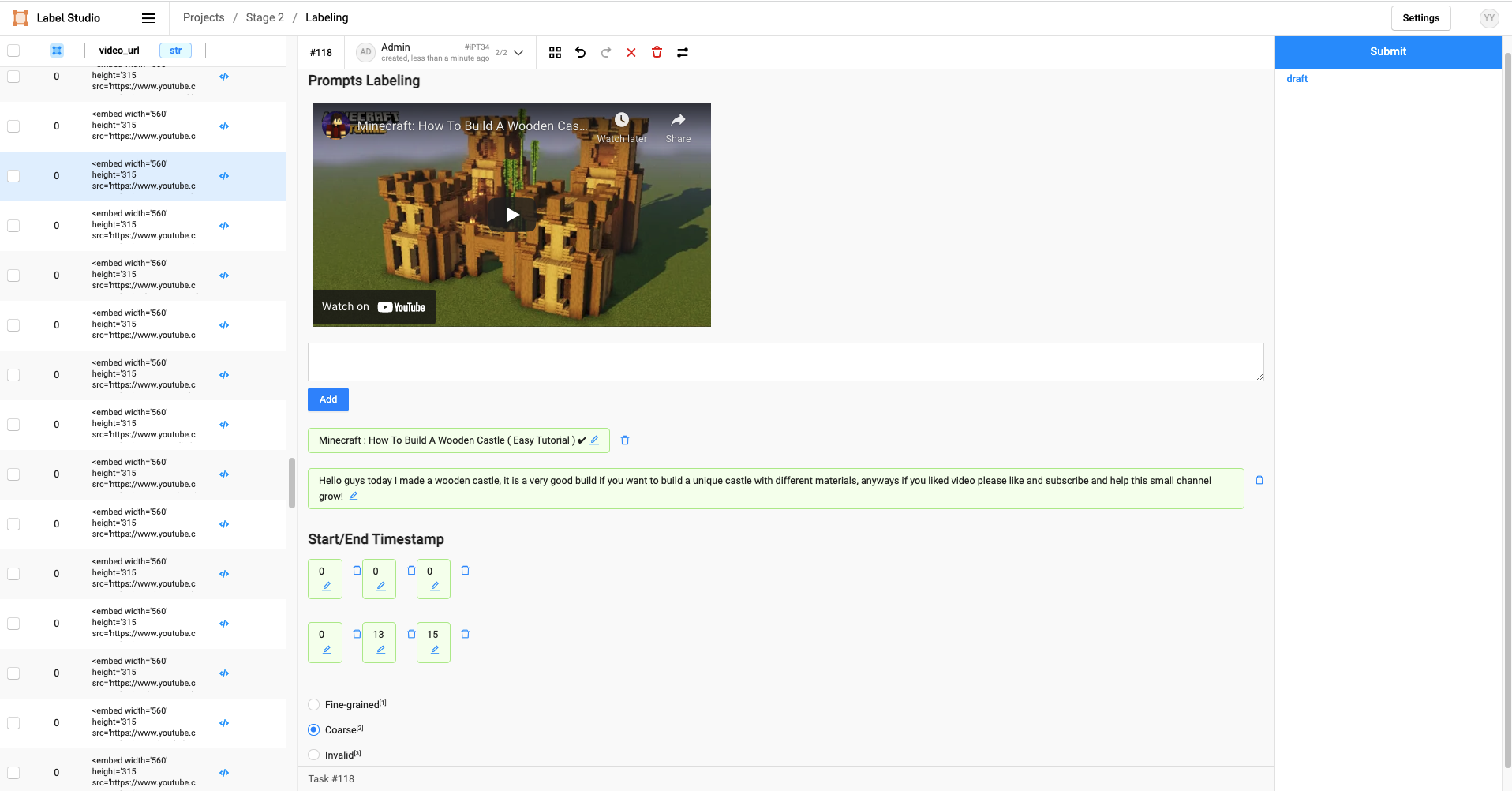}
    \vspace{0.02in}
    \caption{Labeling UI to mine tasks from YouTube. A human annotator can choose to reject the video (\textit{Invalid}), adjust the timestamps, select the title, or edit and expand the original description to be the new task goal.}
    \label{supp:fig:label_ui}
\end{figure}
Our YouTube dataset serves the dual purpose of a rich task source, as many human players demonstrate and narrate creative missions in the tutorial playlists. To collect high-quality tasks and accompanying videos, we design a 3-stage pipeline that makes it easy to find and annotate interesting tasks. 
\begin{enumerate}[label=Stage \arabic*:]
    \item We search for YouTube playlists with the key phrases, ``Minecraft Tutorial'' and ``Minecraft Guide''. Then we apply heuristic rules (see Sec.~\ref{supp:sec:dataset-youtube}) to filter out low-quality videos;
    \item We only show the title of the video to a human annotator through a command-line interface, who makes a binary decision to accept or reject it as a potential task. This step is typically very fast, taking a few seconds on average;
    \item For the accepted tasks in stage 2, we design a labeling UI using Label Studio~\cite{labelstudio} that displays the full video and YouTube description. A human annotator can choose to reject the video, adjust the timestamps, select the title, or refine the description to be the task goal (Fig.~\ref{supp:fig:label_ui}). Through this pipeline, we extract 1,042 task ideas from the common wisdom of a huge number of veteran Minecraft gamers. Some examples are ``\textit{make an automated mining machine}'' and ``\textit{grow cactus up to the sky}''.  
\end{enumerate}

\subsection{GPT-3 Guidance}

We leverage OpenAI's \texttt{GPT-3-davinci} API to automatically generate detailed guidance for a subset of the tasks. Inspired by \cite{kojima2022large}, we adopt the following template to prompt GPT-3: \texttt{How to \{task goal\} in Minecraft? Let's think step by step.} Here are some examples:

The guidance for the task ``\textit{find material and craft a gold pickaxe}'' is \texttt{
1) Find a place with a lot of trees; 
2) Cut down the trees and gather the wood; 
3) Find a place with a lot of stone; 
4) Mine the stone and gather the cobblestone; 
5) Find a place with a lot of iron; 
6) Mine the iron and gather the iron ingots;
7) Find a place with a lot of gold;
8) Mine the gold and gather the gold ingots; 
9) Craft a gold pickaxe.
}

The guidance for the task ``\textit{sail on boat with a sheep}'' is \texttt{
1) Find a boat;
2) Place the sheep in the boat;
3) Right-click on the boat with an empty hand to get in;
4) Use the WASD keys to move the boat. The sheep should stay in the boat.
}

\subsection{Playthrough: Defeat the Ender Dragon}

Our benchmarking suite includes a special task called ``Playthrough''. The agent is initialized bare-handed in a freshly created world and aims to defeat the \textit{Ender dragon}, which is considered the final boss of Minecraft.
This task holds a unique position in our benchmark because killing the dragon means ``beating the game'' in the traditional sense of the phrase, and is considered the most significant achievement for a new player. This boss is optional and plenty of people choose to skip it without affecting their open-ended game experience.

``Playthrough'' is technically a programmatic task, because we can check the simulator state for the defeat of the Ender dragon. However, we decide to create its own category due to the uniqueness as well as the sheer difficulty of the task. The mission requires lots of preparation, exploration, agility, and trial-and-error, which may take a regular human player many days to complete. It would be extremely long horizon (hundreds of thousands of steps) and difficult for an agent to tackle. We consider this one of the moonshot goals in \minedojo.

\section{Internet-Scale Database}
\label{supp:sec:dataset}

We upload our databases to \texttt{\href{https://zenodo.org}{zenodo.org}}, which is an open repository platform operated by CERN. The data DOIs, URLs, and licenses are listed below. In this section, we describe our database properties and data collection process in details.

\vspace{0.1in}
\resizebox{1.0\textwidth}{!}{%
\begin{tabular}{@{}r|l|l@{}}
\toprule
Database & DOI & License \\ 
\midrule
YouTube & \texttt{\href{https://doi.org/10.5281/zenodo.6641142}{10.5281/zenodo.6641142}} & Creative Commons Attribution 4.0 International (CC BY 4.0) \\
Wiki & \texttt{\href{https://doi.org/10.5281/zenodo.6640448}{10.5281/zenodo.6640448}} & Creative Commons Attribution Non Commercial Share Alike 3.0 Unported \\
Reddit & \texttt{\href{https://doi.org/10.5281/zenodo.6641114}{10.5281/zenodo.6641114}} & Creative Commons Attribution 4.0 International (CC BY 4.0) \\
\bottomrule
\end{tabular}
}
\vspace{0.1in}

\subsection{YouTube Videos and Transcripts}
\label{supp:sec:dataset-youtube}
\begin{figure}
    \centering
    \includegraphics[width=.6\linewidth]{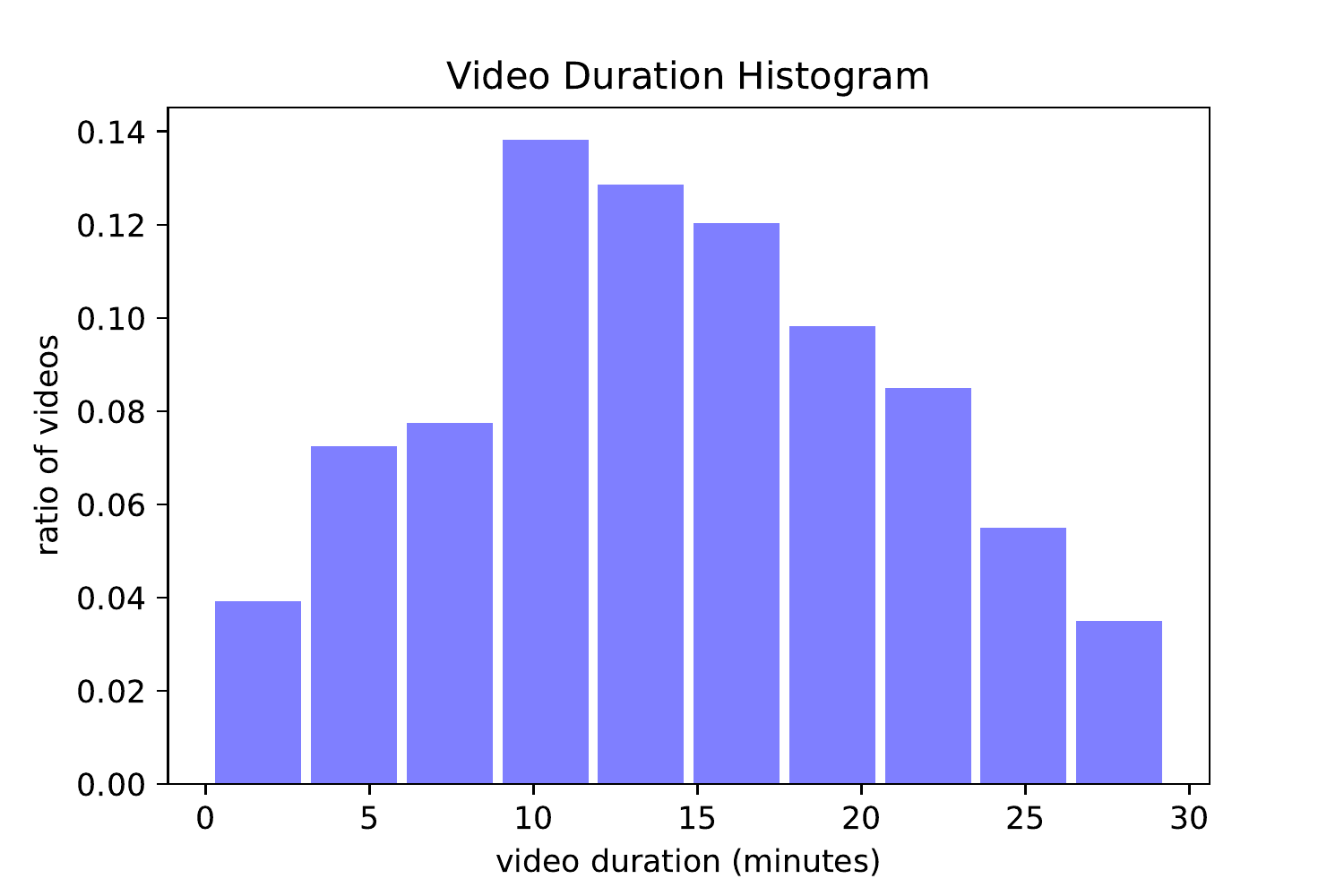}
    \caption{Distribution of YouTube video duration. The histogram is trimmed by the 85th percentile to hide much longer videos that can run for many hours.}
    \label{supp:fig:youtube_dist}
\end{figure}

Minecraft is among the most streamed games on YouTube~\cite{gerblick_2021}. Human players have demonstrated a stunning range of creative activities and sophisticated missions that take hours to complete. We collect 33 years worth of video and 2.2B words in the accompanying English transcripts. The distribution of video duration is shown in Fig.~\ref{supp:fig:youtube_dist}.
The time-aligned transcripts enable the agent to ground free-form natural language in video pixels and learn the semantics of diverse activities without laborious human labeling.

We use the official YouTube Data API~\cite{youtubeapi} to collect our database, following the procedure below:
\begin{enumerate}[label=\alph*)]
    \item Search for \textit{channels} that contain Minecraft videos using a list of keywords, e.g., ``Minecraft'', ``Minecraft Guide'', ``Minecraft Walkthrough'', ``Minecraft Beginner''. We do not directly search for videos at this step because there is a limit of total results returned by the API;
    \item Search for all the video IDs uploaded by each channel that we obtain at the previous step. There are many false positives at this step because some channels (like gaming news channel) may cover a range of topics other than Minecraft;
    \item To remove the false positives, we rely on the video category chosen by the user when the video was uploaded and filter out all the videos that do not belong to the Minecraft category;
    \item To curate a language-grounded dataset, we favor videos that have English transcripts, which can be manually uploaded by the user, automatically transcribed from audio, or automatically translated from another language by the YouTube engine. For each video, we filter it out if 1) the view count is less than 100; or 2) the aspect ratio is less 1; or 3) the duration is less than 1 minute long; or 4) marked as age-restricted.
    \item To further clean the dataset and remove potentially harmful contents, we employ the Detoxify~\cite{detoxify} tool to process each video title and description. Detoxify is trained on Wikipedia comments to predict multiple types of toxicity like threats, obscenity, insults, and identity-based hate speech. We delete a video if the toxicity probability in any category is above 0.5.
\end{enumerate}

We release all the video IDs along with metadata such as video titles, view counts, like counts, duration, and FPS. In line with prior practice~\cite{kay2017kinetics}, we do not release the actual MP4 files and transcripts due to legal concerns.

\subsection{Minecraft Wiki}
\label{supp:sec:dataset-wiki}
\begin{figure}
    \centering
    \includegraphics[width=\linewidth]{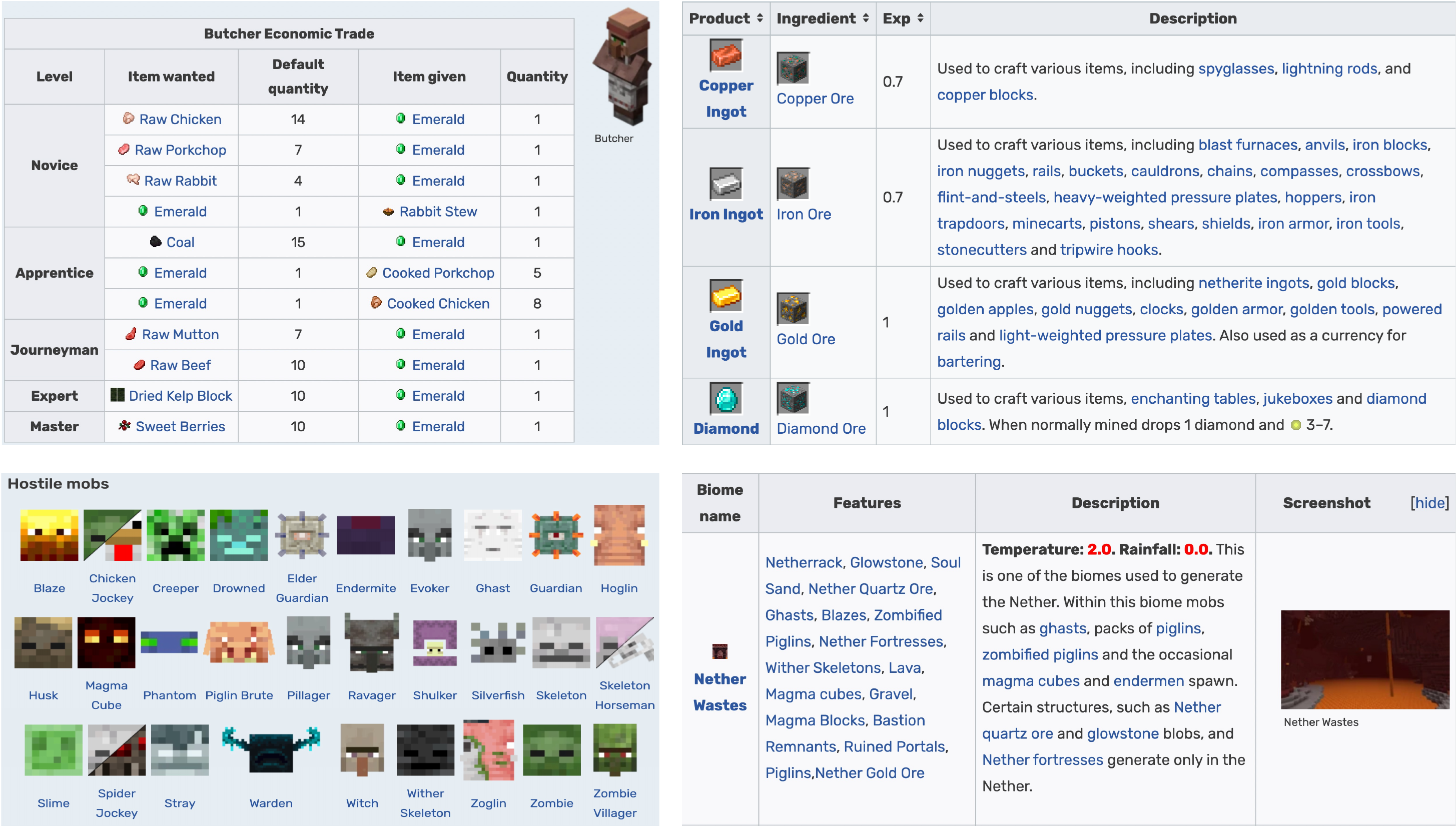}
    \caption{Wiki dataset examples. Closewise order: Villager trade table, mineral ingredient descriptions, monster gallery, and terrain explanation.}
    \label{supp:fig:wiki_supp}
\end{figure}
\begin{figure}
    \centering
    \includegraphics[width=\linewidth]{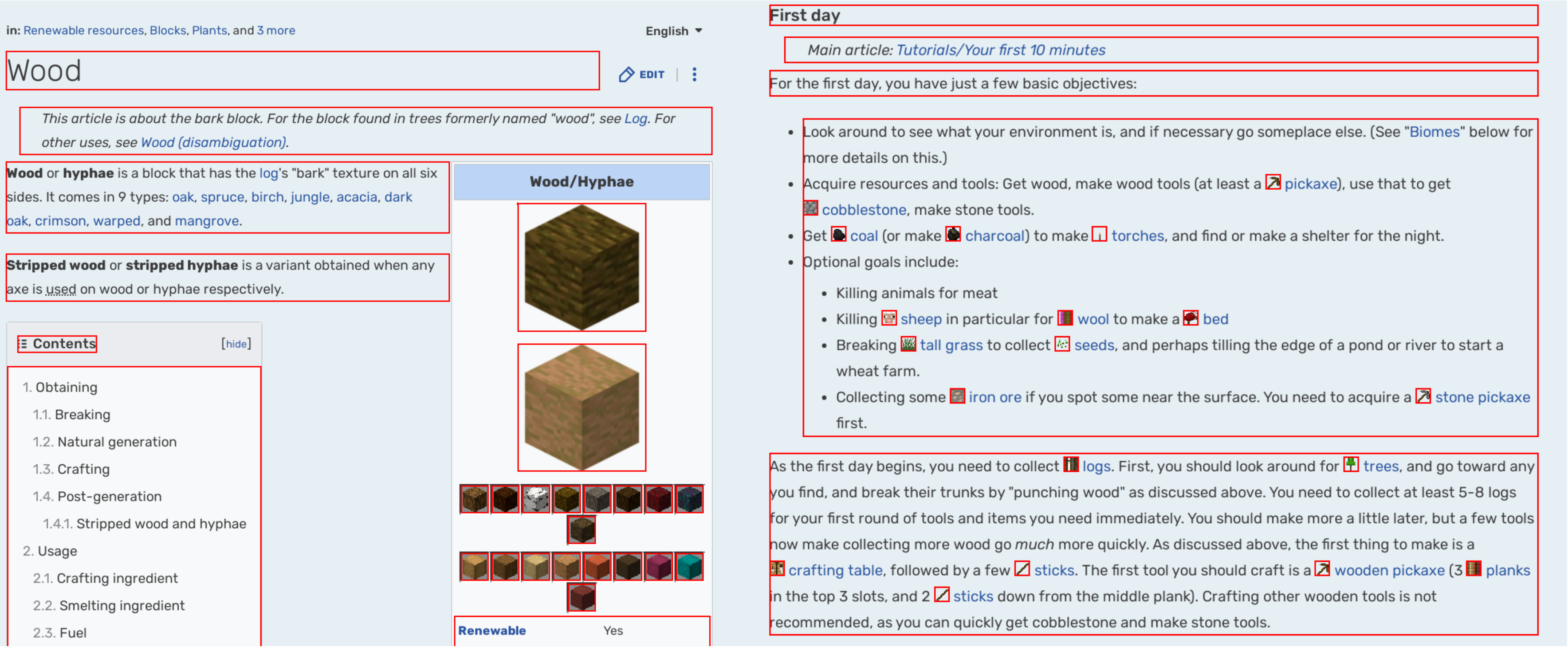}
    \caption{\Rebuttal{More Wiki database examples with bounding boxes (annotated in red). Left: wood block introduction; right: first day tutorial.}}
    \label{supp:fig:wiki_bbox}
\end{figure}

The Wiki pages cover almost every aspect of the game mechanics, and supply a rich source of unstructured knowledge in multimodal tables, recipes, illustrations, and step-by-step tutorials (example screenshots in Fig.~\ref{supp:fig:wiki_supp} \Rebuttal{and Fig.~\mbox{\ref{supp:fig:wiki_bbox}}}). We use Selenium~\cite{selenium} to scrape 6,735 pages that interleave text, images, tables, and diagrams. 
We elaborate the details of each web element scraped by Selenium:

\begin{enumerate}[label=\alph*)]
    \item \textbf{Screenshot.} Using Selenium's built-in function, we take a full screenshot of the rendered Wiki page in order to preserve the human-readable visual formatting. We also record the bounding boxes of each salient web element on the page. 
    \item \textbf{Text.} We hand-select several HTML tags that likely contain meaningful text data, such as \texttt{p, h1, h2, ul, dl}.
    \item \textbf{Images and Animations.} We download the raw source file of each image element (JPG, PNG, GIF, etc.), as well as the corresponding caption if available. There are also animation effects enabled by JavaScript on the Wiki. We save all image frames in the animation. 
    \item \textbf{Sprites.} Sprite elements are micro-sized image icons that are typically embedded in text to create multimodal tutorials and explanations. We save all the sprites and locate their bounding boxes within the text too. 
    \item \textbf{Tables.} We save the text content and bounding box of each cell that a table element contains. We store the header cells separately as they carry the semantic meaning of each column. A table can be easily reconstructed with the stored text strings and bounding boxes. 
\end{enumerate}

\subsection{Reddit}
\label{supp:sec:dataset-reddit}
\begin{figure}[t]
    \vspace{-0.2in}
    \centering
    \includegraphics[width=.6\linewidth]{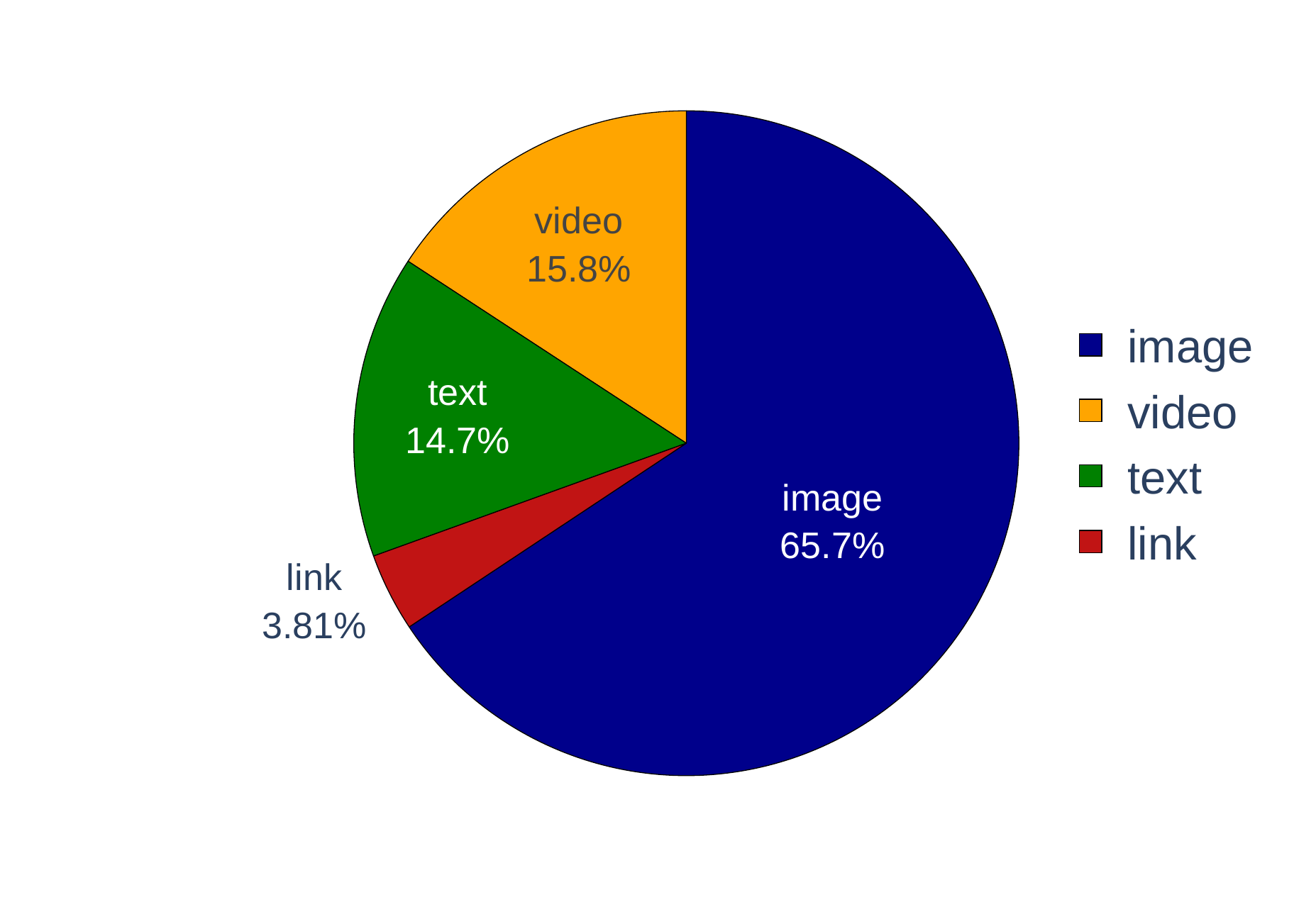}
    \vspace{-0.2in}
    \caption{Distribution of Reddit post types.}
    \label{supp:fig:reddit_dist}
\end{figure}
\begin{figure}
    \centering
    \includegraphics[width=\linewidth]{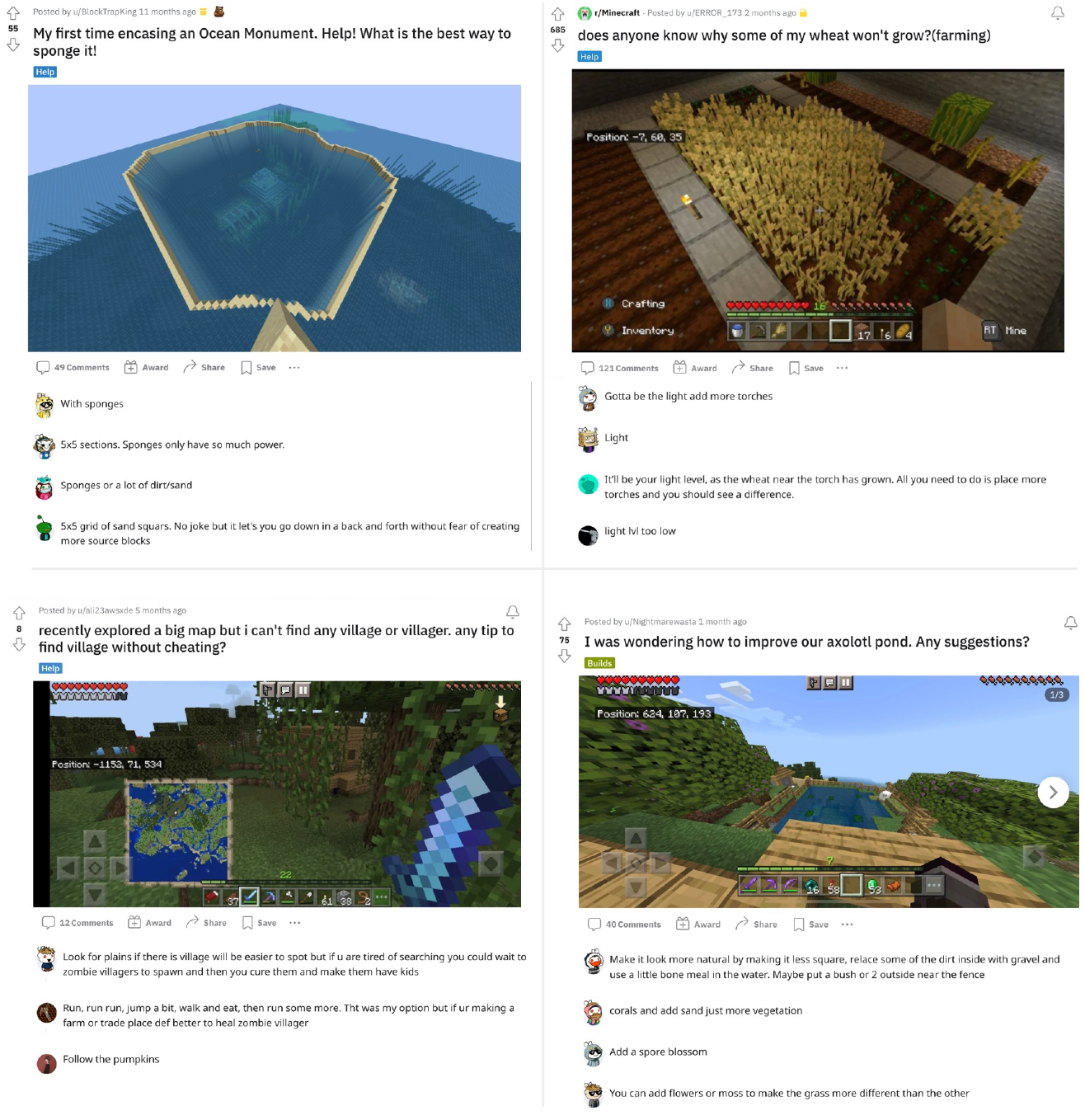}
    \caption{Examples of posts and comment threads from the Reddit database.}
    \label{supp:fig:reddit_supp}
\end{figure}
 There are more than 1M subreddits (i.e., Reddit topics) where people can discuss a wide range of themes and subjects. Prior works use Reddit data for conversational response selection~\cite{al-rfou2016conversational, yang-etal-2018-learning, henderson2019repository} and abstractive summarization~\cite{volske-etal-2017-tl, DBLP:conf/naacl/KimKK19}.
 The \href{https://www.reddit.com/r/minecraft}{r/Minecraft} subreddit contains free-form discussions of game strategies and images/videos showcases of Minecraft builds and creations (examples in Fig.~\ref{supp:fig:reddit_supp}).
The distribution of post types is shown in Fig.~\ref{supp:fig:reddit_dist}. 

To scrape the Reddit contents, we use PRAW~\cite{praw}, a Python wrapper on top of the official Reddit API. Our procedure is as follows:

\begin{enumerate}[label=\alph*)]
    \item Obtain the ID and metadata (e.g. post title, number of comments, content, score) of every post in the ``r/Minecraft'' subreddit since it was created. For quality control, we only consider posts with scores (upvotes) $\geq 5$ and not marked as NSFW.
    \item Determine each post's type. There are 4 native post types - text, image/video, link, and poll. We group text and poll posts together as text posts, and store their body text. 
    For image/video and link posts, we store the source file URLs on external media hosting sites like Imgur and Gfycat. Based on the URL of each link post, we classify it as an image post, a video post or a general link post.
    \item Scrape the comments and store the parent ID of each comment so that we can reconstruct the threaded discussion. 
    \item Similarly to our YouTube database (Sec.~\ref{supp:sec:dataset-youtube}), we run  Detoxify~\cite{detoxify} on the scraped Reddit contents to filter out potentially toxic and harmful posts.
\end{enumerate}

We release all post IDs and their corresponding metadata. We also provide a Python function based on PRAW for researchers to download the post contents after obtaining a license key for the official Reddit API.

\section{\mineclip Algorithm Details}
\label{supp:sec:mineclip}

We implement all our neural networks in PyTorch v1.11 \cite{paszke2019pytorch}. Training \mineclip uses the PyTorch-Lightning framework \cite{falcon2019pytorchlightning}, pre-trained models hosted on HuggingFace \cite{wolf2019huggingfaces}, and the \texttt{x-transformers} library for Transformer variants \cite{wang2021xtransformers}.

\subsection{Video-Text Pair Extraction}

Similar to VideoCLIP~\cite{xu2021videoclip}, we sample 640K pairs of 16-second video snippets and time-aligned English transcripts by the following procedure:

\begin{enumerate}[label=\arabic*)]
    \item Collect a list of keywords corresponding to the supported entities, blocks, and items in Minecraft;
    \item Perform string matching over our YouTube video transcripts to obtain 640K text segments;
    \item For each matched transcript segment, randomly grow it to 16 $\sim$ 77 tokens (limited by CLIP's context length);
    \item Randomly sample a timestamp within the start and end time of the matched transcript as the center for the video clip;
    \item Randomly grow the video clip from the center timestamp to 8 $\sim$ 16 seconds. 
\end{enumerate}

\subsection{Architecture} 
\mineclip architecture is composed of three parts:

\para{Frame-wise image encoder $\phi_I$} We use the ViT-B/16 architecture \cite{dosovitskiy2020image} to compute a 512-D embedding for each RGB frame. We initialize the weights from OpenAI CLIP's public checkpoint \cite{radford2021clip} and only finetune the last two layers during training. The input resolution is  $160 \times 256$, which is different from CLIP's default $224 \times 224$ resolution. We adapt the positional embeddings via bicubic interpolation, which does not introduce any new learnable parameters. 

\para{Temporal aggregator $\phi_a$} Given a sequence of frame-wise RGB features, a temporal aggregator network summarizes the sequence into one video embedding. After the aggregator, we insert two extra layers of residual CLIP Adapter \cite{gao2021clipadapter}. The residual weight is initialized such that it is very close to an identity function at the beginning of training. We consider two variants of $\phi_a$:

\begin{enumerate}
    \item Average pooling (\mineclipavg): a simple, parameter-free operator. It is fast to execute but loses the temporal information, because average pooling is permutation-invariant. 
    \item Self-Attention (\mineclipattn): a 2-layer transformer encoder with 512 embedding size, 8 attention heads, and Gated Linear Unit variant with Swish activation \cite{shazeer2020glu,chowdhery2022palm}. The transformer sequence encoder is relatively slower, but captures more temporal information and achieves better performance in our experiments (Table~\ref{table:core-experiments}).
\end{enumerate}

\begin{table*}
\caption{Training hyperparameters for \mineclip. }
\label{supp:table:hyperparams_mineclip}
\vskip 0.1in
\centering
\resizebox{0.55\textwidth}{!}{%
\begin{tabular}{@{}r|l@{}}
\toprule
Hyperparameter & Value \\ 
\midrule
LR schedule & Cosine with warmup \cite{loshchilov17cosinelr} \\
Warmup steps & 500 \\
Peak LR & 1.5e-4 \\
Final LR & 1e-5 \\
Weight decay & 0.2 \\
Layerwise LR decay & 0.65 \\
Pre-trained layers LR multiplier & $0.5 \times$ \\
Batch size per GPU & 64 \\
Parallel GPUs & 8 \\
Video resolution & $160 \times 256$ \\
Number of frames & 16 \\
Image encoder & \texttt{ViT-B/16} \cite{dosovitskiy2020image} \\

\bottomrule
\end{tabular}
}
\end{table*}

\para{Text encoder $\phi_G$} We use a 12-layer 512-wide GPT model with 8 attention heads \cite{radford2018gpt,radford2019gpt2}. The input string is converted to lower-case byte pair encoding with a 49,152 vocabulary size, and capped at 77 tokens. We exactly follow the text encoder settings in CLIP and initialize the weights from their public checkpoint. Only the last two layers of $\phi_G$ is finetuned during training.  

\subsection{Training}

We train \mineclip on the 640K video-text pairs for 2 epochs. 
We sample 16 RGB frames from each video uniformly, and apply temporally-consistent random resized crop \cite{carreira2017i3d,fan2020rubiks} as data augmentation. 
We use Cosine learning rate annealing with 500 gradient steps of warming up \cite{loshchilov17cosinelr}. We apply a lower learning rate  ($\times 0.5$) on the pre-trained weights and layer-wise learning rate decay for better finetuning \cite{he2021mae}. Training is performed on 1 node of $8 \times$ V100 GPUs with FP16 mixed precision \cite{micikevicius2018mixedprecision} via the PyTorch native \texttt{amp} module. All hyperparameters are listed in Table \ref{supp:table:hyperparams_mineclip}.

\section{Policy Learning Details}
\label{supp:sec:rl}
\begin{algorithm}[t!]
	\caption{PPO-SI Interleaved Training}\label{alg:ppo_si}
	\KwIn{policy $\pi_\theta$, value function $VF(\cdot)$, SI buffer threshold $\Delta$, SI frequency $\omega$}
	Initialize empty SI buffers for all tasks $\mathbf{D}_{SI} \leftarrow \{ \emptyset, \forall T \in \text{training tasks}\}$\;
	Initialize a counter for simulator steps $counter \leftarrow 0$\;
	\While(){not done}{
	    Collect set of trajectories for all tasks $\{\tau_T, \forall T \in \text{training tasks} \}$ by running policy $\pi_\theta$ in (parallel) environments\;
	    \ForAll(){$\mathcal{D}_{SI, T}$}{
	      \uIf{$\tau_T$ is successful}{
	        $\mathcal{D}_{SI, T} \leftarrow \mathcal{D}_{SI, T} \cup \tau_T$
	      }
	      \uElseIf{$\tau_T$'s episode return $\geq \mu_{\text{return}}(\mathcal{D}_{SI, T}) + \Delta \times \sigma_{\text{return}}(\mathcal{D}_{SI, T})$}{
    	      $\mathcal{D}_{SI, T} \leftarrow \mathcal{D}_{SI, T} \cup \tau_T$
	      }
	    }
	    Increase $counter$ accordingly\;
	    Update $\pi_\theta$ following Equation~\ref{eq:ppo_objective}\;
	    Fit $VF(\cdot)$ by regression on mean-squared error\;
	    \uIf{$\mathbbm{1}(counter \bmod \omega = 0)$}{
	        Determine the number of trajectories to sample from each buffer $\#_{\text{sample}} = \min(\{ \vert \mathcal{D}_{SI, T} \vert, \forall T \in \text{training tasks}\})$\;
	        Sample $\#_{\text{sample}}$ trajectories from each buffer in a prioritized manner to construct $\mathcal{D}_{SI}$\;
	        Update $\pi_\theta$ on $\mathcal{D}_{SI}$ with supervised objective\;
	    }
	}
\end{algorithm}

In this section, we elaborate how a trained \mineclip can be adapted as a reward function with two different formulations. We then discuss the algorithm for policy learning. Finally, we demonstrate how we combine self imitation learning and on-policy learning to further improve sample efficiency. 

\subsection{Adapt \mineclip as Reward Function}
\label{supp:sec:rl_mineclip_to_reward}
We investigate two ways to convert \mineclip output to scalar reward, dubbed \direct and \rdelta. The ablation results for \texttt{Animal-Zoo} task group are presented in Table \ref{supp:table:ablation_reward_formulation}.

\para{Direct.} For a task $T$ with the goal description $G$, \mineclip outputs the probability $P_G$ that the observation video semantically corresponds to $G$, against a set of negative goal descriptions $\mathcal{G}^-$. Note that we omit timestep subscript for simplicity. As an example, for the task ``shear sheep'', $G$ is ``shear a sheep'' and $\mathcal{G}^-$ may include negative prompts like ``milk a cow'', ``hunt a sheep'', ``hunt a cow'', etc. 
To compute the \direct reward, we further process the raw probability using the formula  $r = \max( P_G - \frac{1}{N_T}, 0)$ where $N_T$ is the number of prompts passed to \mineclip. $\frac{1}{N_T}$ is the baseline probability of randomly guessing which text string corresponds to the video. We threshold $r$ at zero to avoid highly uncertain probability estimates below the random baseline. We call the variant without the post-processing \direct-Naive: $r = P_G$ as the reward signal for every time step.

\para{Delta.} The \direct formulation yields strong performance when the task is concerned with moving creatures, e.g. farm animals and monsters that run around constantly. However, we discover that \direct is suboptimal if the task deals with static objects, e.g., ``find a nether portal''. Simply using the raw probability from \mineclip as reward can cause the learned agent to stare at the object of interest but fail to move closer and interact. Therefore, we propose to use an alternative formulation, \rdelta, to remedy this issue. Concretely, the reward value at timestep $t$ becomes $r_t = P_{G, t} - P_{G, t - 1}$. We empirically validate that this formulation provides better shaped reward for the task group with static entities.

\begin{table*}
\caption{Ablation on different \mineclip reward formulations.}
\label{supp:table:ablation_reward_formulation}
\vskip 0.1in
\centering
\begin{tabular}{C{0.1\textwidth}|r|cccccc}

\toprule
Group & Tasks & \direct & \direct-Naive & \rdelta \\ \midrule 

\multirow{4}{*}{\addRoboFig{task1}} & Milk Cow & $\bestscore{64.5 \pm 37.1}$ & $8.6 \pm 1.2$ & $7.6 \pm 5.2$ \\
 & Hunt Cow & $\bestscore{83.5 \pm 7.1\hphantom{0}}$ & $0.0 \pm 0.0$ & $0.0 \pm 0.0$ \\
 & Shear Sheep & $\bestscore{12.1 \pm 9.1\hphantom{0}}$ & $0.8 \pm 0.6$ & $1.8 \pm 1.5$ \\
 & Hunt Sheep & $\bestscore{8.1 \pm 4.1}$ & $0.1 \pm 0.2$ & $0.0 \pm 0.0$ \\
\bottomrule

\end{tabular}
\end{table*}

\subsection{Policy Network Architecture}
Our policy architecture consists of three parts: an input feature encoder, a policy head, and a value function. 
To handle multimodal observations (Sec.~\ref{supp:sec:sim_obs_space}), the feature extractor contains several modality-specific components: 

\begin{itemize}
    \item RGB frame: we use the frozen frame-wise image encoder $\phi_I$ in \mineclip to optimize for compute efficiency and provide the agent with good visual representations from the beginning (Sec.~\ref{sec:method-rl}).
    \item Task goal: $\phi_G$ computes the text embedding of the natural language task goal. 
    \item Yaw and Pitch: compute $\sin(\cdot)$ and $\cos(\cdot)$ features respectively, then pass through an MLP. 
    \item GPS: normalize and featurize via MLP.
    \item Voxel: to process the $3 \times 3 \times 3$ surrounding voxels, we embed discrete block names to dense vectors, flatten them, and pass through an MLP.
    \item Past action: our agent is conditioned on its immediate past action, which is embedded and featurized by MLP. 
\end{itemize}

Features from all modalities are concatenated, passed through another fusion MLP, and finally fed into the policy head and value function head. 
We use an MLP to model the policy head that maps from the input feature vectors to the action probability distribution. We use another MLP to estimate the value function, conditioned on the same input features. 

\subsection{RL Training}
\label{sec:appendix_rl_training}

\para{PPO.} We use the popular PPO algorithm ~\cite{schulman2017proximal} (Proximal Policy Optimization) as our RL training backbone. PPO is an on-policy method that optimizes for a surrogate objective while ensuring that the deviation from the previous policy is relatively small. PPO updates the policy network by
\begin{equation}
    \underset{\theta}{\text{maximize}}\;\mathbb{E}_{s,a\sim\pi_{\theta_{\text{old}}}} L(s, a, \theta_\text{old}, \theta),
\end{equation} where
\begin{equation}
    L(s, a, \theta_\text{old}, \theta) = \min \left(\frac{\pi_{\theta}(a|s)}{\pi_{\theta_\text{old}}(a|s)} A^{\pi_{\theta_\text{old}}}(s, a), \text{clip}\left(\frac{\pi_{\theta}(a|s)}{\pi_{\theta_\text{old}}(a|s)}, 1-\epsilon, 1+\epsilon\right)A^{\pi_{\theta_\text{old}}}(s, a)\right).
    \label{eq:ppo_objective}
\end{equation}
$A$ is an estimator of the advantage function (GAE~\cite{DBLP:journals/corr/SchulmanMLJA15} in our case) and $\epsilon$ is a hyperparameter that controls the deviation between the new policy and the old one.

\para{Self Imitation Learning.} We apply self-imitation learning \cite{oh2018self} (SI) to further improve sample efficiency because computing the reward with \mineclip in the loop makes the training more expensive. Self-imitation learning is essentially supervised learning on a buffer $\mathcal{D}_{SI}$ of good trajectories generated by the agent's past self. In our case, the trajectories are generated by the behavior policy during PPO rollouts, and only added to $\mathcal{D}_{SI}$ if it is a \emph{successful} trial or if the episodic return exceeds a certain threshold. Self imitation optimizes $\pi_\theta$ for the objective  
$\mathcal{J}_{SI} = \mathbb{E}_{s, a \sim \mathcal{D}_{SI}} \log \pi_\theta (a \vert s)$ 
with respect to $\theta$.

We alternate between the PPO phase and the SI phase. 
A pseudocode of our interleaved training procedure is given in Algorithm~\ref{alg:ppo_si}. We use a \emph{prioritized} strategy to sample trajectories from the buffer $\mathcal{D}_{SI}$. Specifically, we assign equal probability to all successful trajectories. Unsuccessful trajectories can still be sampled but with lower probabilities proportional to their episodic returns. 

\begin{figure*}[t!]
     \centering
     \begin{subfigure}[b]{0.49\linewidth}
         \centering
         \includegraphics[width=\textwidth]{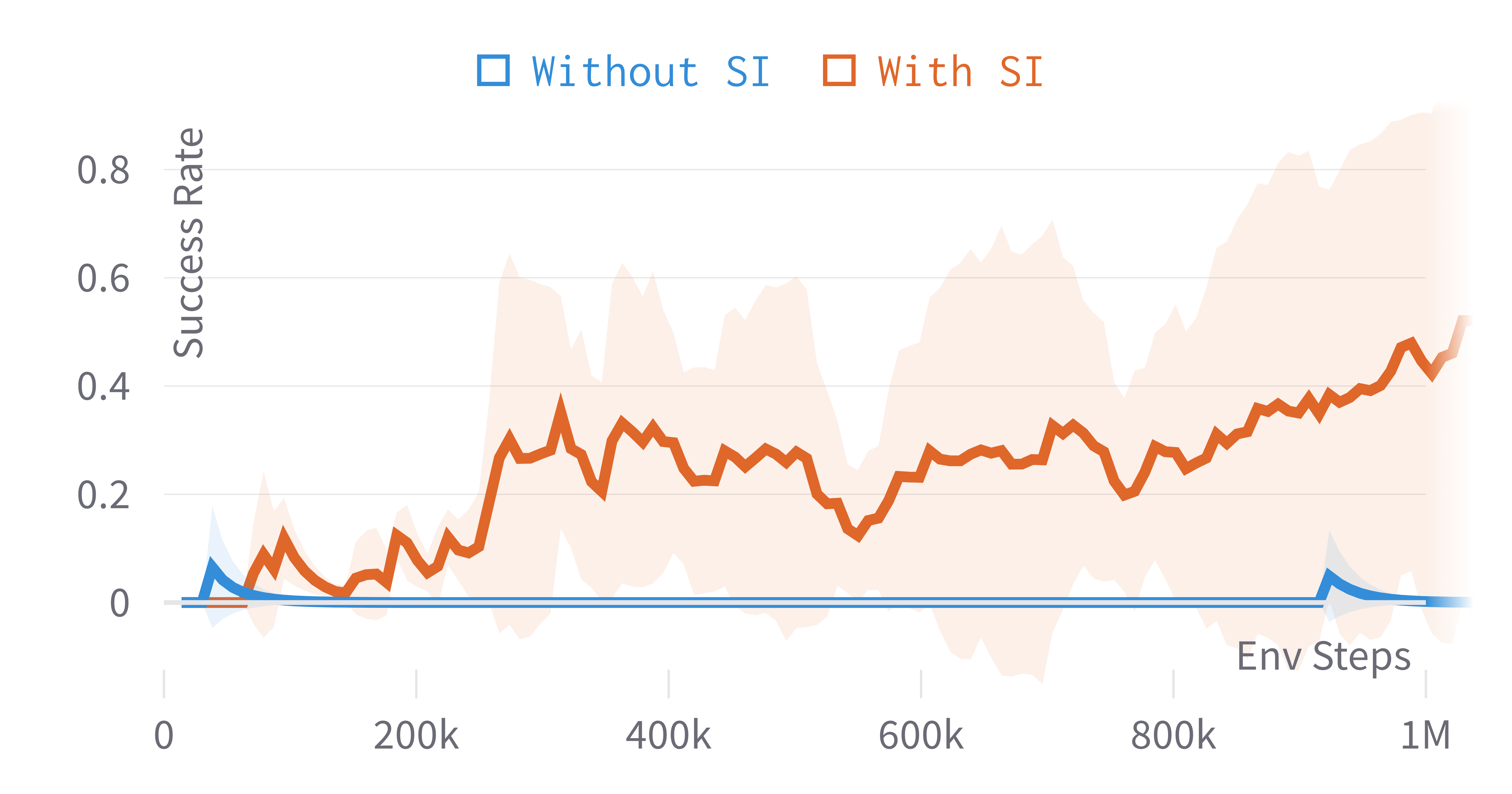}
         \caption{``Milk Cow''}
         \label{fig:si_ablation_milk_cow}
     \end{subfigure}
     \hfill
     \begin{subfigure}[b]{0.49\linewidth}
         \centering
         \includegraphics[width=\textwidth]{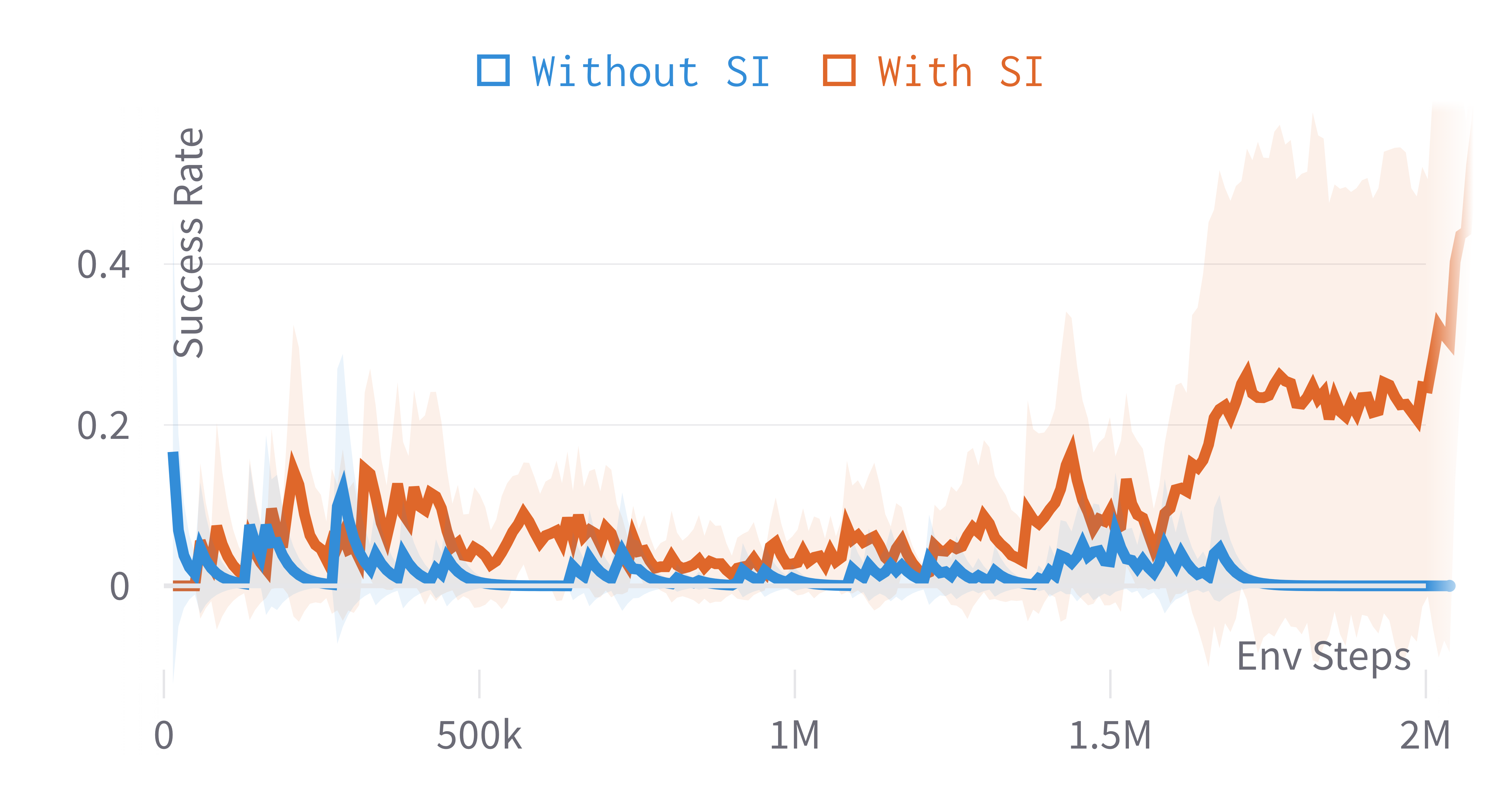}
         \caption{``Shear Sheep''}
         \label{fig:si_ablation_shear_sheep}
     \end{subfigure}
    \caption{Adding the self imitation technique \cite{oh2018self} significantly improves the performance of RL training in \minedojo.}
    \label{supp:fig:ablation_si}
\end{figure*}

In Fig. \ref{supp:fig:ablation_si}, we demonstrate that adding self-imitation dramatically improves the stability, performance, and sample efficiency of RL training in \minedojo.  

\section{Experiment Details}
\label{supp:sec:experiments}

\subsection{Task Details}
\label{supp:sec:task_details}
We experiment with three task groups with four tasks per group. We train one multi-task agent for each group. In this section, we describe each task goals, initial setup, and the manual dense-shaping reward function.

\para{\underline{Animal Zoo}:} 4 Programmatic tasks on hunting or harvesting resource from animals. We spawn various animal types (pig, sheep, and cow) in the same environment to serve as distractors. It is considered a failure if the agent does not take action on the correct animal specified by the prompt.

\begin{itemize}
    \item  \textbf{Milk Cow}: find and approach a cow, then obtain milk from it with an empty bucket. The prompt is \texttt{milk a cow}. We initialize the agent with an empty bucket to collect milk. We also spawn sheep, cow, and pig nearby the agent. The manual dense reward shaping is a navigation reward based on geodesic distance obtained from privileged LIDAR. The combined reward passed to PPO can be formulated as $r_t = \lambda_{\text{nav}}\max(d_{\text{min}, t - 1} - d_{\text{min}, t}, 0) + \lambda_{\text{success}} \mathbbm{1}(\text{milk collected})$, where $\lambda_{\text{nav}} = 10$ and $\lambda_{\text{success}} = 200$. $d_{\text{min}, t} = \min(d_{\text{min}}, d_{t})$ where $d_{\text{min}}$ denotes the minimal distance to the cow that the agent has achieved so far in the episode history. 
    \item \textbf{Hunt Cow}: find and approach a cow, then hunt with a sword. The cow will run away so the agent needs to chase after it. The prompt is \texttt{hunt a cow}. We initialize the agent with a diamond sword. The manual dense reward shaping consists of two parts, a valid attack reward and a navigation reward based on geodesic distance obtained from privileged LIDAR. Mathematically, the reward is $r_t = \lambda_{\text{attack}}\mathbbm{1}(\text{valid attack}) + \lambda_{\text{nav}}\max(d_{\text{min}, t - 1} - d_{\text{min}, t}, 0) + \lambda_{\text{success}} \mathbbm{1}(\text{cow hunted})$, where $\lambda_{\text{attack}} = 5$, $\lambda_{\text{nav}} = 1$, and $\lambda_{\text{success}} = 200$. We additionally reset $d_\text{min}$ every time the agent hits the cow to encourage the chasing behavior. 
    \item \textbf{Shear Sheep}: find and approach a sheep, then collect wool from the sheep with a shear. The prompt is \texttt{shear a sheep}. We initialize the agent with a shear. The manual dense reward shaping is a navigation reward based on geodesic distance obtained from the privileged LIDAR sensor, similar to ``Milk Cow''.
    \item \textbf{Hunt Sheep}: find and approach a sheep, then hunt with a sword. The sheep will run away so the agent needs to chase after it. An episode will terminate once any entity is hunted. The prompt is \texttt{hunt a sheep}. We initialize the agent with a diamond sword. The manual dense reward shaping consists of two parts, a valid attack reward and a navigation reward based on geodesic distance obtained from the privileged LIDAR sensor, similar to ``Hunt Cow''.
\end{itemize}

\para{\underline{Mob Combat}:} fight 4 different types of hostile monsters: Spider, Zombie, Zombie Pigman (a creature in the \textit{Nether} world), and Enderman (a creature in the \textit{End} world). The prompt template is \texttt{"Combat \{monster\}"}.
For all tasks within this group, we initialize the agent with a diamond sword, a shield, and a full suite of diamond armors. The agent is spawned in the \textit{Nether} for Zombie Pigman task, and in the \textit{End} for Enderman.
The manual dense-shaping reward can be expressed as $r_t = \lambda_{\text{attack}} \mathbbm{1}(\text{valid attack}) + \lambda_{\text{success}} \mathbbm{1}(\texttt{\{monster\}}\text{ hunted})$ where $\lambda_{\text{attack}} = 5$ and $\lambda_{\text{success}} = 200$.

\para{\underline{Creative}:} 4 tasks that do not have manual dense reward shaping or code-defined success criterion.

\begin{itemize}
    \item \textbf{Find Nether Portal}: find and move close to a Nether Portal, then enter the \textit{Nether} world through the portal. The prompt is \texttt{find a nether portal}. 
    \item \textbf{Find Ocean}: find and move close to an ocean. The prompt is \texttt{find an ocean}. 
    \item \textbf{Dig Hole}: dig holes in the ground. The prompt is \texttt{dig a hole}. We initialize the agent with an iron shovel.
    \item \textbf{Lay Carpet}: lay down carpets to cover the wooden floor inside a house. The prompt is \texttt{put carpets on the floor}. We initialize the agent with a number of carpets in its inventory. 
\end{itemize}

\Rebuttal{Note that we categorize ``Find Nether Portal'' and ``Find Ocean'' as Creative tasks even though they seem similar to object navigation \mbox{\cite{batra2020objectnav}}. While finding terrains and other structures is semantically well defined, it is not easy to define a function to evaluate success automatically because the simulator does not have the exact location information of these structures given a randomly generated world. In principle, we can make a sweep by querying each chunk of voxels in the world to recognize the terrains, but that would be prohibitively expensive. Therefore, we opt to use MineCLIP as the reward signal and treat these tasks as Creative.} 

\begin{table*}[t!]
\caption{Hyperparameters in RL experiments. ``\texttt{\{state\}} MLP'' refers to MLPs to process observations of compass, GPS, and voxel blocks. ``Embed Dim'' denotes the same dimension size used to embed all discrete observations into dense vectors.}
\label{supp:table:hyperparams_rl}

\centering
\resizebox{\textwidth}{!}{%
\begin{tabular}{rrll|rccc}
\toprule
\multicolumn{4}{c|}{\multirow{2}{*}{NN Architecture}}       & \multicolumn{4}{c}{Training}                                                                                                         \\ \cline{5-8} 
\multicolumn{4}{c|}{}                                       & Hyperparameter              & \multicolumn{1}{c|}{\texttt{Animal-Zoo}}         & \multicolumn{1}{c|}{\texttt{Mob-Combat}}         & \texttt{Creative}           \\ \hline
RGB Feature Size                & \multicolumn{3}{c|}{512}  & Learning Rate               & \multicolumn{1}{c|}{$10^{-4}$}          & \multicolumn{1}{c|}{$10^{-4}$}          & $10^{-4}$          \\
Task Prompt Feature Size        & \multicolumn{3}{c|}{512}  & Cosine Decay Minimal LR     & \multicolumn{1}{c|}{$5 \times 10^{-6}$} & \multicolumn{1}{c|}{$5 \times 10^{-6}$} & $5 \times 10^{-6}$ \\
\texttt{\{state\}} MLP Hidden Size           & \multicolumn{3}{c|}{128}  & $\gamma$                    & \multicolumn{1}{c|}{$0.99$}             & \multicolumn{1}{c|}{$0.99$}             & $0.99$             \\
\texttt{\{state\}} MLP Output Size           & \multicolumn{3}{c|}{128}  & Entropy Weight (Stage 1)    & \multicolumn{1}{c|}{$5 \times 10^{-3}$} & \multicolumn{1}{c|}{$5 \times 10^{-3}$} & $5 \times 10^{-3}$ \\
\texttt{\{state\}} MLP Hidden Depth          & \multicolumn{3}{c|}{2}    & Entropy Weight (Stage 2)    & \multicolumn{1}{c|}{$10^{-2}$}          & \multicolumn{1}{c|}{N/A}                & $10^{-2}$          \\
Embed Dim                       & \multicolumn{3}{c|}{8}    & PPO Optimizer               & \multicolumn{1}{c|}{Adam}               & \multicolumn{1}{c|}{Adam}               & Adam               \\
Num Feature Fusion Layers       & \multicolumn{3}{c|}{1}    & SI Learning Rate            & \multicolumn{1}{c|}{$10^{-4}$}          & \multicolumn{1}{c|}{$10^{-4}$}          & $10^{-4}$          \\
Feature Fusion Output Size      & \multicolumn{3}{c|}{512}  & SI Cosine Decay Minimal LR  & \multicolumn{1}{c|}{$10^{-6}$}          & \multicolumn{1}{c|}{$10^{-6}$}          & $10^{-6}$          \\
Prev Action Conditioning        & \multicolumn{3}{c|}{True} & SI Epoch                    & \multicolumn{1}{c|}{10}                 & \multicolumn{1}{c|}{10}                 & 10                 \\
Policy Head Hidden Size         & \multicolumn{3}{c|}{256}  & SI Frequency (Env Steps)    & \multicolumn{1}{c|}{100K}               & \multicolumn{1}{c|}{100K}               & 100K               \\
Policy Head Hidden Depth        & \multicolumn{3}{c|}{3}    & SI Optimizer                & \multicolumn{1}{c|}{Adam}               & \multicolumn{1}{c|}{Adam}               & Adam               \\
VF Hidden Size        & \multicolumn{3}{c|}{256}  & SI Buffer Threshold         & \multicolumn{1}{c|}{$2 \sigma$}         & \multicolumn{1}{c|}{$2 \sigma$}         & $0.5 \sigma$       \\
VF Hidden Depth        & \multicolumn{3}{c|}{3}    & PPO Buffer Size             & \multicolumn{1}{c|}{100K}               & \multicolumn{1}{c|}{100K}               & 100K               \\
                                & \multicolumn{3}{c|}{}     & Frame Stack                 & \multicolumn{1}{c|}{1}                  & \multicolumn{1}{c|}{1}                  & 1                  \\
                                & \multicolumn{3}{c|}{}     & VF Loss Weight              & \multicolumn{1}{c|}{0.5}                & \multicolumn{1}{c|}{0.5}                & 0.5                \\
                                & \multicolumn{3}{c|}{}     & GAE $\lambda$               & \multicolumn{1}{c|}{0.95}               & \multicolumn{1}{c|}{0.95}               & 0.95               \\
                                & \multicolumn{3}{c|}{}     & Gradient Clip               & \multicolumn{1}{c|}{10}                 & \multicolumn{1}{c|}{10}                 & 10                 \\
                                & \multicolumn{3}{c|}{}     & PPO $\epsilon$              & \multicolumn{1}{c|}{0.2}                & \multicolumn{1}{c|}{0.2}                & 0.2                \\
                                & \multicolumn{3}{c|}{}     & Action Smooth Weight        & \multicolumn{1}{c|}{$10^{-7}$}          & \multicolumn{1}{c|}{$10^{-7}$}          & $10^{-7}$          \\
                                & \multicolumn{3}{c|}{}     & Action Smooth Window Size   & \multicolumn{1}{c|}{3}                  & \multicolumn{1}{c|}{3}                  & 3                  \\
                                & \multicolumn{3}{c|}{}     & \mineclip Reward Formulation & \multicolumn{1}{c|}{\direct}     & \multicolumn{1}{c|}{\direct}     & \rdelta              \\ \bottomrule
\end{tabular}

}
\end{table*}

\subsection{Observation and Action Space}

We use a subset of the full observation and action space listed in Sec.~\ref{supp:sec:sim_obs_space} and \ref{supp:sec:sim_action_space}, because the tasks in our current experiments do not involve actions like crafting or inventory management. Our observation space consists of RGB frame, compass, GPS, and Voxels.

Our action space is a trimmed version of the full action space. It consists of movement control, camera control, ``use'' action, and ``attack'' action, which add up to 89 discrete choices. Concretely, it includes 81 actions for discrete camera control ($9 \times 9$ resulted from the Cartesian product between yaw and pitch, each ranges from $-60$ degree to $60$ degree with a discrete interval of $15$ degree). It also includes 6 movement actions (forward, forward + jump, jump, back, move left, and move right) and 2 functional actions of ``use'' and ``attack''. Note that the ``no-op'' action is merged into the 81 camera actions.

\subsection{RL Training}
All hyperparameters used in our RL experiment are listed in Table~\ref{supp:table:hyperparams_rl}. We visualize the learned behaviors of 4 tasks in Figure~\ref{fig:behavior_visualization}. Demos of more tasks can be found on our website \weburl.

\para{Action smoothing.} 
Due to the stochastic nature of PPO, we observe a lot of action jittering in the agent's behavior during training. This leads to two negative effects that degrade the learning performance: 1) exploration difficulty due to inconsistent action sequence. For example, the agent may be required to take multiple consecutive \texttt{attack} actions in order to complete certain tasks; and 2) rapidly switching different movement and camera motions result in videos that are highly non-smooth and disorienting. This causes a domain gap from the training data of \mineclip, which are typically smooth human gameplay videos. Therefore, the reward signal quality deteriorates significantly.

To remedy the issue, we impose an action smoothing loss to be jointly optimized with the PPO objective (Eq.~\ref{eq:ppo_objective}) during training. Concretely, consider a sliding window $\mathcal{W}$ with window size $\vert \mathcal{W} \vert$ that contains $\vert \mathcal{W} \vert$ consecutive action distributions $\mathcal{W} = \{\pi_{t - \vert \mathcal{W} \vert + 1}, \pi_{t - \vert \mathcal{W} \vert + 2}, \ldots, \pi_{t} \}$, the action smoothing loss is defined as 
\begin{equation}
    \mathcal{L}_{\text{smooth}} = \frac{1}{\vert \mathcal{W} \vert} \sum_{i = 1}^{\vert \mathcal{W} \vert - 1} KL(\pi_t \Vert \pi_{t - \vert \mathcal{W} \vert + i}),
\end{equation}
where $KL(\cdot)$ denotes Kullback–Leibler divergence. 
 
\para{Multi-stage training for multi-task RL.} 
Due to hardware limitations, we are not able to run a large number of parallel simulators for all tasks in a task group. Therefore, we adopt a multi-stage strategy to split the tasks and train them sequentially with a single policy network. For the task groups \texttt{Animal-Zoo} and \texttt{Creative}, we split the four tasks into two stages of two parallel training tasks each. We carry over the self-imitation buffers when switching to the next stage. We also follow the recommended practice in \cite{nikishin2022primacy} and reset the policy head at the beginning of stage 2 to encourage exploration and reduce overfitting. We adopt a similar replay buffer balancing strategy as \cite{gupta2022metamorph} to prevent any task from dominating the training. 

\subsection{Evaluation}
\label{supp:sec:mineclip_complex_creative}
In this section, we elaborate on our human and automatic evaluation procedure for Creative tasks.
We first ask the human annotators to manually label 100 successful and 100 failure trajectories. This produces a combined dataset of 200 trajectories with groundtruth binary labels to evaluate the learned reward functions.  
On this dataset, we run \mineclip to produce step-wise rewards and compute a score that averages over each trajectory. We then apply K-means clustering with $K = 2$ to all scores and determine a decision boundary $\delta$ from the mean of the two centroids. A trajectory with a score greater than $\delta$ is classified as successful, and vice versa for failure. In this way, we essentially convert \mineclip to a binary classifier. 
The quality of \mineclip can be measured by the F1 score of its binary classification output against the human labels. We demonstrate that \mineclip has high agreements with humans (Table \ref{table:eval-metric-quality}), and thus qualifies as an effective automatic evaluation metric for Creative tasks in the absence of human judges. 

To further investigate \mineclip{}'s evaluation on more complex Creative tasks,
we annotate 50 YouTube video segments each for 5 more tasks that are much more semantically complex: ``build a farm'', `` build a fence'', ``build a house'', ``ride a minecart'', and ``build a swimming pool''. We then run \mineclip evaluation on these videos against a negative set. As shown in Table \ref{table:supp-mineclip-eval-more}, though not perfect, \mineclip generally has a positive agreement with human judgment. We note that the current \mineclip is a proof-of-concept step in leveraging internet data for automated evaluation, and further scaling on more training data and parameters may lead to more improvements. Meanwhile, human judgment remains a useful and important alternative \cite{openai2022dalle2,saharia2022imagen}.

\begin{table}[t!]
\caption{
\Rebuttal{\mbox{\mineclip{}'s} evaluation on more complex Creative tasks. 
Numbers represent F1 scores between \mbox{\mineclip}'s evaluation on tasks success and human labels. Scaled to percentage for better readability.}
}
\label{table:supp-mineclip-eval-more}
\vskip 0.1in
\centering

\resizebox{1\textwidth}{!}{
\begin{tabular}{r|ccccccc}
\toprule
 Tasks & Build a Farm & Build a Fence & Build a House & Ride a Minecart & Build a Swimming Pool  \\ \midrule 

 Ours (Attn) & $\bestscore{78.7}$ & $\bestscore{91.4}$ & $\bestscore{63.7}$ & $95.9$ & $85.0$\\

 Ours (Avg) & $73.4$ & $83.1$ & $37.4$ & $\bestscore{96.9}$ & $\bestscore{94.7}$\\
 
 \openaiclip & $62.5$ & $24.5$ & $52.9$ & $70.0$ & $71.7$\\

\bottomrule
\end{tabular}
}
\end{table}

\section{Limitations and Potential Societal Impact}

Unlike human demonstrations \cite{vinyals2019alphastar} or offline RL datasets \cite{fu2020d4rl}, our YouTube dataset contains only the video screen observations but not the actual control actions. This allows us to scale up the dataset tremendously, but at the same time poses a challenge to imitation learning algorithms that require observation-action pairs to learn. 
Our proposed algorithm, \mineclip, side-steps this problem by learning a reward model, but we believe that directly inferring the human expert policy from YouTube is another important direction complementary to our approach. There are promising techniques that can potentially overcome this limitation, such as the Learning-from-Observation (LfO) family of algorithms \cite{torabi2019lfosurvey,torabi2018behavioral,stadie2017thirdperson,edwards2018imitatinglatent}. 

Our database is scraped from the internet, which inevitably contains offensive YouTube videos or toxic Reddit posts. While we have made our best effort to filter out these harmful contents (Sec.~\ref{supp:sec:dataset-youtube}), there can still be undesirable biases and toxicity that elude our automatic filters. Furthermore, we advocate the use of large pre-trained language models in our main paper, and \mineclip is finetuned from the pre-trained weights of OpenAI CLIP \cite{radford2021clip}. These foundation models are known to contain harmful stereotypes and generate hateful commentary \cite{brown2020gpt3,bommasani2021foundation,gehman2020realtoxicityprompts}. We ask the researchers who will use our code and database to exercise their best judgment during new model development to avoid any negative social impact.   

\section{Datasheet}

We present a Datasheet \cite{datasheet21gebru} for documentation and responsible usage of our internet knowledge databases.

\subsection{Motivation}

\para{For what purpose was the dataset created?} We create this internet-scale multimodal knowledge base to facilitate research towards open-ended, generally capable embodied agents.

\para{Who created the dataset (e.g., which team, research group) and on
behalf of which entity (e.g., company, institution, organization)?} This knowledge base was created by Linxi Fan (Nvidia), Guanzhi Wang (Caltech), Yunfan Jiang (Stanford), Ajay Mandlekar (Nvidia), Yuncong Yang (Columbia), Haoyi Zhu (SJTU), Andrew Tang (Columbia), De-An Huang (Nvidia), Yuke Zhu (Nvidia and UT Austin), and Anima Anandkumar (Nvidia and Caltech).

\subsection{Distribution}
\para{Will the dataset be distributed to third parties outside of the entity (e.g., company, institution, organization) on behalf of which the dataset was created?} Yes, the dataset is publicly available on the internet.

\para{How will the dataset will be distributed (e.g., tarball on website, API, GitHub)?}
All datasets can be downloaded from \url{https://zenodo.org/}. Please refer to this table of URL, DOI, and licensing: 

\vspace{0.1in}
\resizebox{1.0\textwidth}{!}{%
\begin{tabular}{@{}r|l|l@{}}
\toprule
Database & DOI & License \\ 
\midrule
YouTube & \texttt{\href{https://doi.org/10.5281/zenodo.6641142}{10.5281/zenodo.6641142}} & Creative Commons Attribution 4.0 International (CC BY 4.0) \\
Wiki & \texttt{\href{https://doi.org/10.5281/zenodo.6640448}{10.5281/zenodo.6640448}} & Creative Commons Attribution Non Commercial Share Alike 3.0 Unported \\
Reddit & \texttt{\href{https://doi.org/10.5281/zenodo.6641114}{10.5281/zenodo.6641114}} & Creative Commons Attribution 4.0 International (CC BY 4.0) \\
\bottomrule
\end{tabular}
}
\vspace{0.1in}

\para{Have any third parties imposed IP-based or other restrictions on the data associated with the instances?} No.

\para{Do any export controls or other regulatory restrictions apply to the dataset or to individual instances?} No.

\subsection{Maintenance}
\para{Who will be supporting/hosting/maintaining the dataset?} The authors will be supporting, hosting, and maintaining the dataset. 

\para{How can the owner/curator/manager of the dataset be contacted (e.g., email address)?} Please contact Linxi Fan (\texttt{linxif@nvidia.com}), Guanzhi Wang (\texttt{guanzhi@caltech.edu}), and Yunfan Jiang (\texttt{yunfanj@cs.stanford.edu}).

\para{Is there an erratum?} No. We will make announcements if there is any. 

\para{Will the dataset be updated (e.g., to correct labeling errors, add new instances, delete instances)?} Yes. New updates will be posted on \weburl.

\para{If the dataset relates to people, are there applicable limits on the retention of the data associated with the instances (e.g., were the individuals in question told that their data would be retained for a fixed period of time and then deleted)?}N/A.

\para{Will older versions of the dataset continue to be supported/hosted/maintained?} Yes, old versions will be permanently accessible on \href{https://zenodo.org}{zenodo.org}.

\para{If others want to extend/augment/build on/contribute to the dataset, is there a mechanism for them to do so?} Yes, please refer to \weburl.

\subsection{Composition}

\para{What do the instances that comprise the dataset represent?} 

For YouTube videos, our data is in JSON format with video URLs and metadata. We do not provide the raw MP4 files for legal concerns. For Wiki, we provide the text, images, tables, and diagrams embedded on the web pages. For Reddit, our data is in JSON format with post IDs and metadata, similar to YouTube. Users can reconstruct the Reddit dataset by running our script after obtaining an official Reddit API license key.    

\para{How many instances are there in total (of each type, if appropriate)?} There are more than 730K YouTube videos with 2.2B words of transcripts, 6,735 Wiki pages with 2.2M bounding boxes of visual elements, and more than 340K Reddit posts with 6.6M comments.

\para{Does the dataset contain all possible instances or is it a sample
(not necessarily random) of instances from a larger set?} We provide all instances in our Zenodo data repositories. 

\para{Is there a label or target associated with each instance?} No.

\para{Is any information missing from individual instances?} No. 

\para{Are relationships between individual instances made explicit
(e.g., users’ movie ratings, social network links)?} We provide metadata for each YouTube video link and Reddit post ID. 

\para{Are there recommended data splits (e.g., training, development/validation,
testing)?} No. The entire database is intended for pre-training.

\para{Are there any errors, sources of noise, or redundancies in the
dataset?} Please refer to Sec.~\ref{supp:sec:dataset}

\para{Is the dataset self-contained, or does it link to or otherwise rely on
external resources (e.g., websites, tweets, other datasets)?} We follow prior works \cite{kay2017kinetics} and only release the video URLs of YouTube videos due to legal concerns. Researchers need to acquire the MP4 and transcript files separately. Similarly, we only release the post IDs for the Minecraft Reddit database, but we also provide a script that can reconstruct the full Reddit dataset given a free official license key. 

\para{Does the dataset contain data that might be considered confidential?} No. 

\para{Does the dataset contain data that, if viewed directly, might be offensive, insulting, threatening, or might otherwise cause anxiety?} We have made our best efforts to detoxify the contents via an automated procedure. Please refer to Sec.~\ref{supp:sec:dataset}.

\subsection{Collection Process}

The collection procedure, preprocessing, and cleaning are explained in details in Sec.~\ref{supp:sec:dataset}.

\para{Who was involved in the data collection process (e.g., students, crowdworkers, contractors) and how were they compensated (e.g., how much were crowdworkers paid)?} All data collection, curation, and filtering are done by \minedojo coauthors. 

\para{Over what timeframe was the data collected?} The data was collected between Dec. 2021 and May 2022.

\subsection{Uses}
\para{Has the dataset been used for any tasks already?} Yes, we have used the \minedojo YouTube database for agent pre-training. Please refer to Sec.~\ref{sec:experiments} and Sec.~\ref{supp:sec:experiments} for algorithmic and training details.

\para{What (other) tasks could the dataset be used for?} Our knowledge base is primarily intended to facilitate research in open-ended, generally capable embodied agents. However, it can also be broadly applicable to research in video understanding, document understanding, language modeling, multimodal learning, and so on. 

\para{Is there anything about the composition of the dataset or the way
it was collected and preprocessed/cleaned/labeled that might impact future uses?}  No.

\para{Are there tasks for which the dataset should not be used?}  We strongly oppose any research that intentionally generates harmful or toxic contents using our YouTube, Wiki, and Reddit data.

\end{document}